\documentclass{article}
\PassOptionsToPackage{table,dvipsnames}{xcolor}
\usepackage[preprint]{colm2026_conference}

\usepackage[T1]{fontenc}
\usepackage[utf8]{inputenc}
\usepackage{microtype}
\usepackage{hyperref}
\usepackage{url}
\usepackage{booktabs}
\usepackage{graphicx}
\usepackage{apple_table_preamble}

\usepackage{amsmath,amssymb}
\usepackage{algorithm}
\usepackage{algpseudocode}
\usepackage{subcaption}
\usepackage{adjustbox}
\usepackage{makecell}
\usepackage[most]{tcolorbox}

\usepackage{lineno}

\graphicspath{{./}}

\definecolor{darkblue}{rgb}{0, 0, 0.5}
\definecolor{linkblue}{HTML}{6155F5}
\hypersetup{
  colorlinks=true,
  citecolor=darkblue,
  linkcolor=darkblue,
  urlcolor=darkblue,
  pdftitle={ClawBench: Can AI Agents Complete Everyday Online Tasks?},
  pdfauthor={Yuxuan Zhang, Yubo Wang, Yipeng Zhu, Penghui Du, Junwen Miao, Xuan Lu, Zhuofeng Li, Xingwei Qu, Zhengkang Guo, Yuanzhe Shen, Dingjie Song, Han Zhou, Tuney Zheng, Xian Wu, Hao Yu, Songcheng Cai, Yi Lu, Yunzhuo Hao, Minyi Lei, Liang Chen, Kai Zou, Huifeng Yin, Wendong Xu, Dongfu Jiang, Ping Nie, Jiaheng Liu, Wenhu Chen, Kelsey R. Allen}
}

\title{ClawBench: Can AI Agents Complete Everyday Online Tasks?}

\makeatletter
\renewcommand{\@maketitle}{
  \begingroup
    \raggedright
    \leftskip=0pt plus 0pt
    {\fontsize{14pt}{17pt}\selectfont\bfseries\color{linkblue} \@title\par}
    \vskip 0.16in
    \@author \par
    \vskip 0.3in
  \endgroup
}
\makeatother

\renewcommand{\headrulewidth}{0pt}

\author{
{\small
\mbox{\textbf{Yuxuan Zhang}\,\textsuperscript{1,2,3,\S}},
\mbox{\textbf{Yubo Wang}\,\textsuperscript{2,5}},
\mbox{\textbf{Yipeng Zhu}\,\textsuperscript{1}},
\mbox{\textbf{Penghui Du}\,\textsuperscript{3}},
\mbox{\textbf{Junwen Miao}\,\textsuperscript{4}},
\mbox{\textbf{Xuan Lu}\,\textsuperscript{6}},\\[2pt]
\mbox{\textbf{Zhuofeng Li}\,\textsuperscript{5}},
\mbox{\textbf{Xingwei Qu}\,\textsuperscript{12}},
\mbox{\textbf{Zhengkang Guo}\,\textsuperscript{13}},
\mbox{\textbf{Yuanzhe Shen}\,\textsuperscript{13}},
\mbox{\textbf{Dingjie Song}\,\textsuperscript{14}},
\mbox{\textbf{Han Zhou}\,\textsuperscript{3}},\\[2pt]
\mbox{\textbf{Tuney Zheng}\,\textsuperscript{5}},
\mbox{\textbf{Xian Wu}\,\textsuperscript{3}},
\mbox{\textbf{Hao Yu}\,\textsuperscript{10}},
\mbox{\textbf{Songcheng Cai}\,\textsuperscript{5}},
\mbox{\textbf{Yi Lu}\,\textsuperscript{5}},
\mbox{\textbf{Yunzhuo Hao}\,\textsuperscript{5}},
\mbox{\textbf{Minyi Lei}\,\textsuperscript{5}},\\[2pt]
\mbox{\textbf{Liang Chen}\,\textsuperscript{7,\textdagger}},
\mbox{\textbf{Kai Zou}\,\textsuperscript{11}},
\mbox{\textbf{Huifeng Yin}\,\textsuperscript{7}},
\mbox{\textbf{Wendong Xu}\,\textsuperscript{7,\S}},
\mbox{\textbf{Dongfu Jiang}\,\textsuperscript{2,5,\textdagger}},
\mbox{\textbf{Ping Nie}\,\textsuperscript{5,\S}},\\[2pt]
\mbox{\textbf{Jiaheng Liu}\,\textsuperscript{15}},
\mbox{\textbf{Wenhu Chen}\,\textsuperscript{2,5,\textdagger}},
\mbox{\textbf{Kelsey R.\ Allen}\,\textsuperscript{1,2,\textdagger}}
}\\[0.45em]
{\scriptsize\linespread{1.25}\selectfont
\strut\textsuperscript{1}University of British Columbia \quad
\textsuperscript{2}Vector Institute \quad
\textsuperscript{3}Etude AI \quad
\textsuperscript{4}Carnegie Mellon University\strut\\
\strut\textsuperscript{5}University of Waterloo \quad
\textsuperscript{6}SJTU \quad
\textsuperscript{7}UniPat AI \quad
\textsuperscript{9}HKUST \quad
\textsuperscript{10}Tsinghua University \quad
\textsuperscript{11}Netmind.ai\strut\\
\strut\textsuperscript{12}University of Manchester \quad
\textsuperscript{13}Fudan University \quad
\textsuperscript{14}Lehigh University \quad
\textsuperscript{15}Nanjing University\strut
}\\[0.15em]
{\normalsize\href{https://claw-bench.com}{\color{linkblue}\texttt{\bfseries https://claw-bench.com}}}
}

\newcommand{\authorrolefootnote}{%
  \par\vspace*{-16pt}\noindent\rule{12pc}{0.4pt}\par
  \vspace{2.6pt}%
  {\footnotesize\textsuperscript{\S}Project Lead. \quad \textsuperscript{\textdagger}Advisors.\par}%
}

\makeatletter
\newcommand{\UseProjectInputs}{\def\input@path{{./}{arr_submit/sec/}}}
\newcommand{\UseLocalInputs}{\def\input@path{{}}}
\makeatother

\begin{document}

\ifcolmsubmission
\linenumbers
\fi

\maketitle

\UseProjectInputs
\vspace*{-2pt}
\begin{figure}[H]
\centering
\captionsetup{font=small,skip=5pt}
\includegraphics[width=\textwidth]{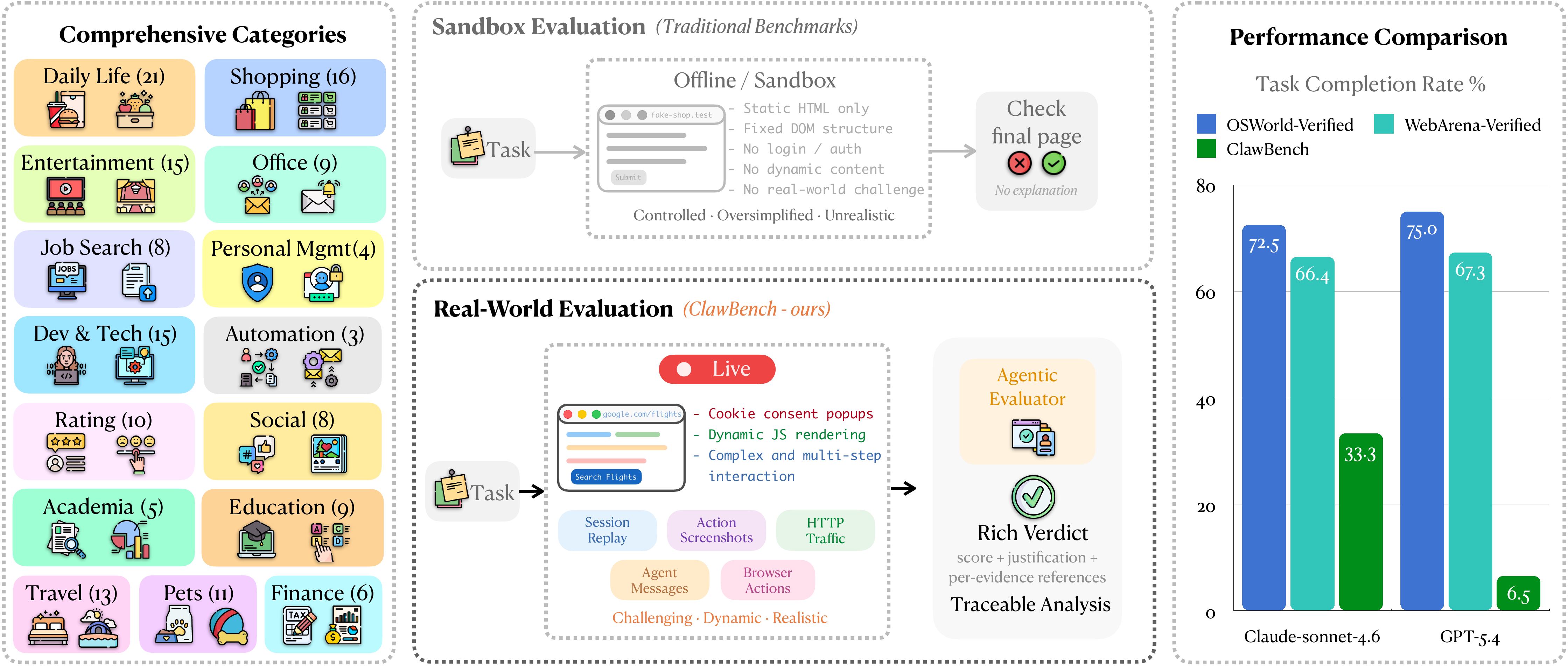}
\caption{\textsc{ClawBench} overview.
\textbf{Left}: \textsc{ClawBench} contains 153 realistic online tasks spanning 15 everyday life categories and 144 live web platforms.
\textbf{Middle}: unlike prior web-agent benchmarks that rely on offline sandbox environments with static pages and limited read-only interactions, \textsc{ClawBench} evaluates agents on live production websites involving dynamic, multi-step, and write-capable interactions, while preventing unintended real-world side effects and supporting fine-grained, step-level traceable evaluation.
\textbf{Right}: despite strong performance on existing benchmarks, frontier models still achieve substantially lower success rates on \textsc{ClawBench}, highlighting the large gap between controlled benchmark performance and real-world online task completion.}
\label{fig:teaser}
\end{figure}
\begin{figure}[H]
\centering
\begin{minipage}[t]{0.42\textwidth}
    \centering
    \includegraphics[width=\linewidth,height=0.23\textheight,keepaspectratio]{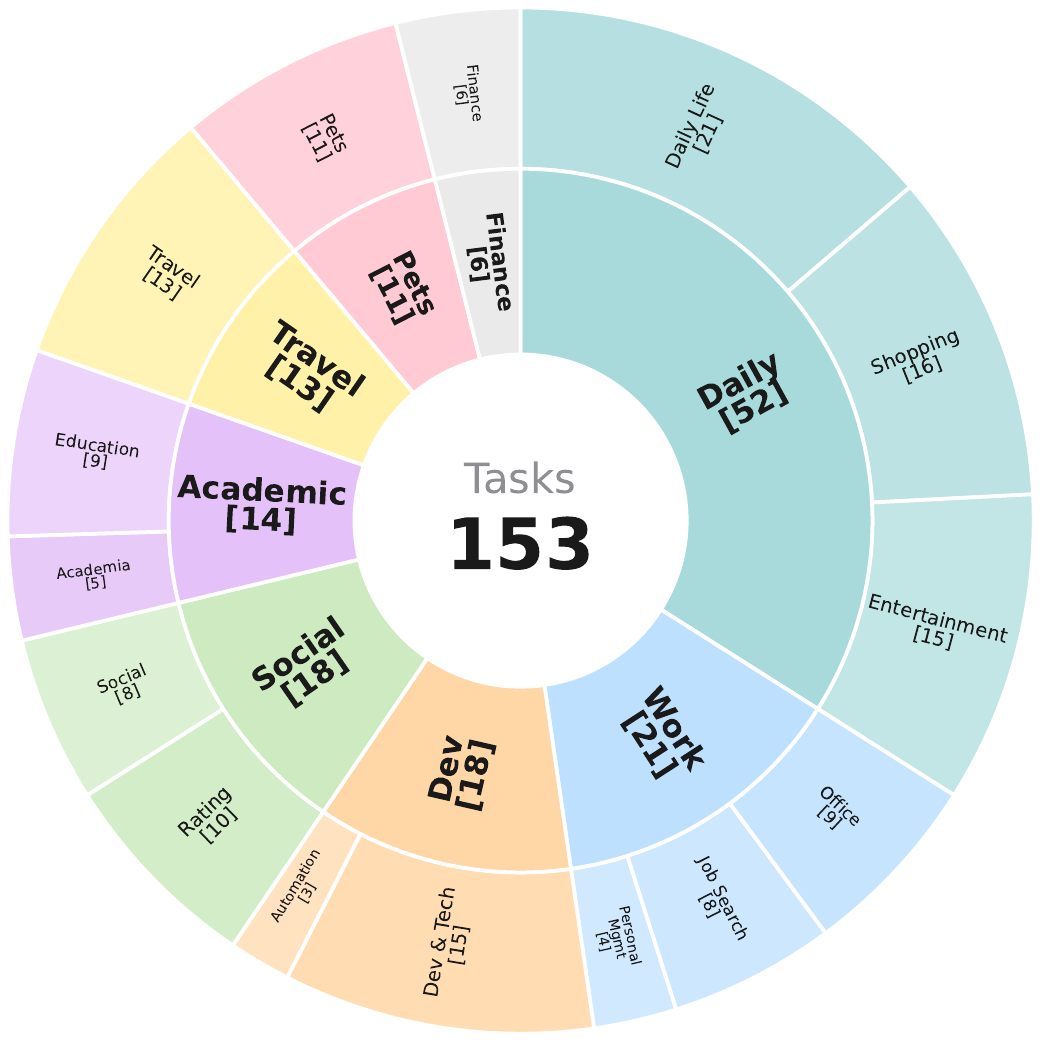}
    \captionof{figure}{Task taxonomy of \textsc{ClawBench}. Inner ring: 8 high-level category groups; outer ring: 15 fine-grained categories.}
    \label{fig:taxonomy}
\end{minipage}
\hfill
\begin{minipage}[t]{0.55\textwidth}
    \centering
    \includegraphics[width=\linewidth,height=0.23\textheight,keepaspectratio]{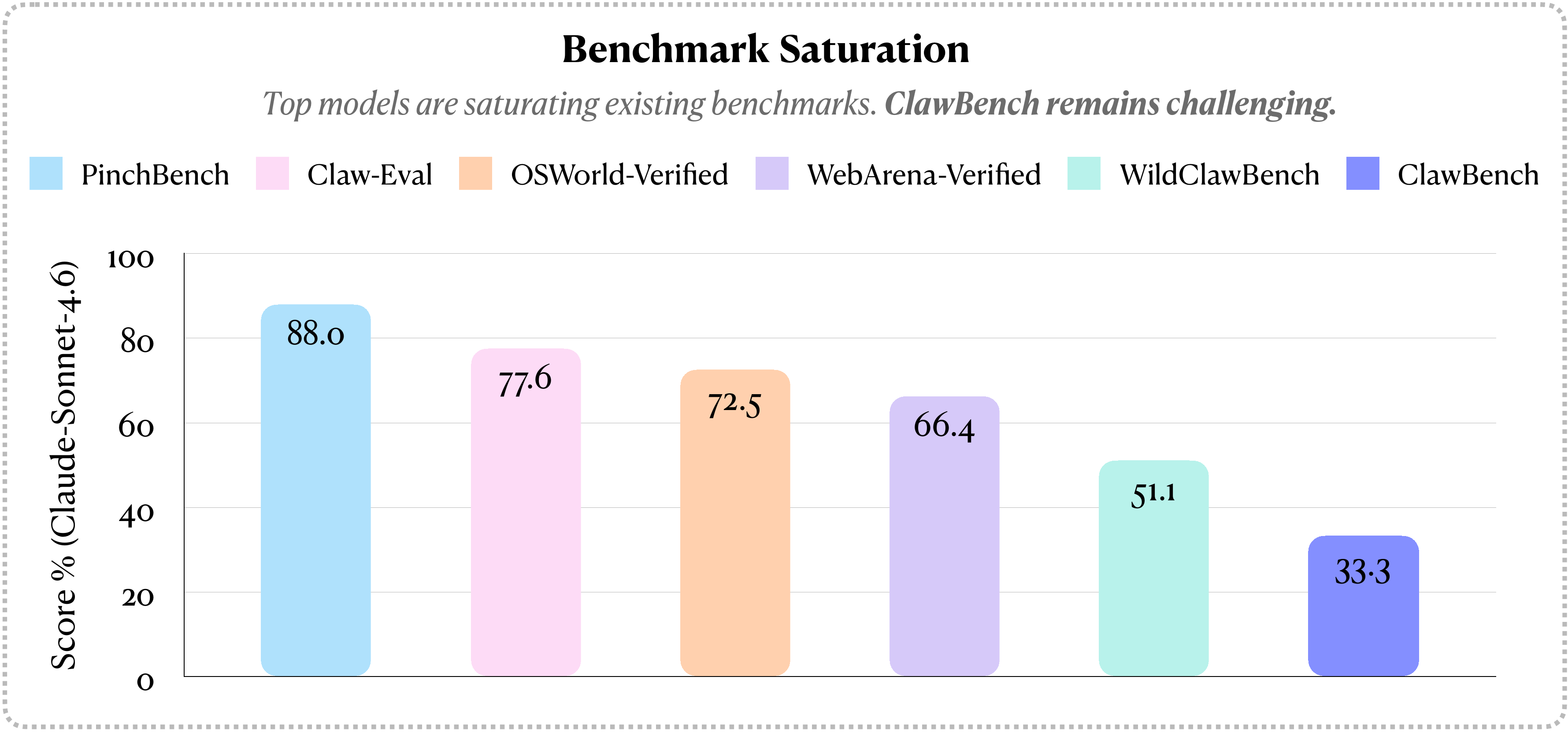}
    \captionof{figure}{Benchmark saturation comparison. Claude-Sonnet-4.6 performs substantially better on existing web-agent benchmarks than on \textsc{ClawBench}.}
\label{fig:benchmark_saturation}
\end{minipage}
\end{figure}
\authorrolefootnote
\clearpage
\begin{abstract}
AI agents may be able to assist with emails and documents, but can they reliably complete everyday online workflows on real websites?
Everyday online tasks offer a realistic yet unsolved testbed
for evaluating the next generation of AI agents.
To this end, we introduce \textsc{ClawBench}, an evaluation framework comprising 153 everyday online tasks that people need to accomplish regularly in their lives and work,
spanning 144 platforms across 15 categories,
from completing purchases and booking appointments
to submitting job applications.
These tasks require capabilities beyond existing benchmarks, such as obtaining relevant information from user-provided documents, navigating multi-step workflows across diverse platforms, and write-heavy operations like filling in many detailed forms correctly.
Unlike existing benchmarks that evaluate agents in offline sandboxes with static pages,
\textsc{ClawBench} operates on production websites, preserving the full complexity, dynamic nature, and interaction challenges of real-world web environments.
An interception layer captures and blocks the final submission request,
ensuring safe evaluation without real-world side effects.
Our evaluations of 8 frontier models show that both proprietary
and open-source models complete only a small portion of these tasks.
For example, Claude Sonnet 4.6 achieves only 33.3\%, which exposes gaps in current AI agents.
Progress on \textsc{ClawBench} brings us closer to AI agents that can function as general-purpose assistants.
\end{abstract}

\section{Introduction}

Existing web-agent benchmarks cover important slices of browser use, but none jointly evaluates \textbf{daily-life}, \textbf{write-heavy} tasks on \textbf{live production websites} with safe, traceable scoring.
Static-trace benchmarks such as Mind2Web~\citep{deng2023mind2web} evaluate action prediction over recorded pages rather than task execution.
Sandboxed or self-hosted benchmarks such as WebArena~\citep{zhou2024webarena}, VisualWebArena~\citep{koh2024visualwebarena}, OSWorld~\citep{xie2024osworld}, and TheAgentCompany~\citep{xu2024theagentcompany} provide controlled environments, but necessarily abstract away the instability and heterogeneity of production websites.
Real-web benchmarks such as WebVoyager~\citep{he2024webvoyager}, AssistantBench~\citep{yoran2024assistantbench}, Online-Mind2Web~\citep{xue2025illusion}, and EconWebArena~\citep{liu2025econwebarena} move evaluation to live websites, but focus primarily on read-only information acquisition or answer extraction rather than state-changing workflows.
Claw-Eval~\citep{ye2026claweval} studies trajectory-aware grading in controlled agent settings, but does not evaluate broad everyday transactions across live platforms.

The missing target is whether AI agents can actually complete \textbf{everyday online tasks} that change, or attempt to change, real web state.
Such tasks include purchases, reservations, applications, account updates, and detailed form submissions, where success depends on navigating real interfaces, satisfying platform-specific validation logic, handling authentication or verification flows, and reaching a terminal action that would modify server-side state.
This setting is central to general online assistance, but it creates a difficult evaluation problem: the benchmark must preserve the real website while preventing orders, applications, reservations, or other irreversible side effects from actually reaching production servers.

We introduce \textsc{ClawBench}, a benchmark designed around three properties missing in combination from prior work.
\textit{(1) Daily-life coverage.} \textsc{ClawBench} contains 153 everyday online tasks across 15 life categories and 144 live platforms, as shown in \autoref{fig:teaser}.
\textit{(2) Write-heavy workflows.} Tasks require state-changing interactions such as purchases, reservations, applications, account updates, and detailed form submissions, rather than only information lookup.
\textit{(3) Live-web execution with safe evaluation.} Agents operate on production websites, preserving dynamic interfaces, authentication flows, platform-specific validation, and anti-bot defenses, while \textsc{ClawBench} blocks the final action that would create real-world side effects.

The key safety mechanism in \textsc{ClawBench} is \textbf{final-request interception}.
For each task, human annotators complete the workflow and identify the terminal HTTP request that would commit the irreversible action.
During agent execution, a Chrome extension records browser actions while a Chrome DevTools Protocol (CDP) based instrumentation server monitors outgoing network traffic, captures the terminal request and its payload, and blocks that request before it reaches the server.
All preceding page loads, searches, authentication steps, JavaScript calls, and form-filling interactions proceed on the website, preserving ecological validity while preventing real-world side effects.

We evaluate each run with an \textbf{Agent-as-Judge} evaluator rather than a URL, DOM, or final-page heuristic.
\textsc{ClawBench} records each run as a \textbf{five-layer trajectory}: full-session video, per-step screenshots, HTTP traffic, agent messages, and low-level browser actions.
Human annotators produce a reference trajectory and intercepted reference payload for every task under the same setup.
Given the task instruction, the human reference, the agent trajectory, and the intercepted payload, the Agent-as-Judge assigns a binary task-level verdict with a structured justification grounded in evidence.
At a high level, a run passes when the agent completes the required workflow and either reaches the intercepted final request with correct or semantically equivalent payload values, or is blocked at an allowed terminal point after completing all required prior steps; it fails when key fields are wrong or missing, the agent stops before attempting the terminal action, or external blockers prevent task completion under the evaluation rubric.
We further audit this protocol on human-adjudicated subsets covering two contrastive models, where Agent-as-Judge reaches $84.97$--$93.46\%$ raw agreement with human verdicts (\autoref{tab:judge_human_agreement}); an expanded four-model breakdown is reported in Appendix~\ref{app:safety-audits}.

\begingroup
\begin{table*}[!t]
\centering
\tablefontsize
\caption[Comparison of \textsc{ClawBench} with existing web agent benchmarks]{
Comparison with current web-agent benchmarks.
\textsc{ClawBench} uniquely combines live production websites, write-heavy tasks, human-authored reference trajectories, and multi-layer behavioral recording.}
\label{tab:benchmark_comparison}
\vspace{4pt}
\renewcommand{\arraystretch}{1.35}
\adjustbox{max width=\textwidth}{
\begin{tabular}{l c r r c c c c}
\toprule
\thead{Benchmark}
  & \thead{Environment}
  & \thead{\# Tasks}
  & \thead{\# Sites}
  & \thead{Task Type}
  & \thead{Verification}
  & \thead{Recording}
  & \thead{Human Traj.} \\
\midrule

\altcolor
Mind2Web~\citep{deng2023mind2web}
  & Offline {\scriptsize(static traces)}
  & 2{,}350
  & 137
  & Read-only
  & Action seq.\ match
  & None
  & Partial {\scriptsize(action seq.)} \\

WebArena~\citep{zhou2024webarena}
  & Sandbox {\scriptsize(self-hosted)}
  & 812
  & 5
  & Mixed
  & Script-based
  & None
  & \xmark \\

\altcolor
VisualWebArena~\citep{koh2024visualwebarena}
  & Sandbox {\scriptsize(self-hosted)}
  & 910
  & 3
  & Mixed
  & Script-based
  & None
  & \xmark \\

OSWorld~\citep{xie2024osworld}
  & Sandbox {\scriptsize(VM)}
  & 369
  & 9
  & Mixed
  & Script + screenshot
  & Screenshot
  & \xmark \\

\altcolor
WebVoyager~\citep{he2024webvoyager}
  & Real Web
  & 643
  & 15
  & Read-only
  & LLM-as-judge
  & Screenshot
  & \xmark \\

TheAgentCompany~\citep{xu2024theagentcompany}
  & Sandbox {\scriptsize(self-hosted)}
  & 175
  & 6
  & Mixed
  & Checkpoint-based
  & None
  & \xmark \\

\altcolor
Online-Mind2Web~\citep{xue2025illusion}
  & Real Web
  & 300
  & 136
  & Read-only
  & Human + LLM judge
  & Screenshot
  & \xmark \\

EconWebArena~\citep{liu2025econwebarena}
  & Real Web
  & 360
  & 82
  & Read-only
  & Exact numeric + URL
  & None
  & \xmark \\

\altcolor
Claw-Eval~\citep{ye2026claweval}
  & Sandbox {\scriptsize(Docker + FastAPI)}
  & 139
  & 15
  & Mixed
  & API state check
  & CLI logs
  & \xmark \\

\midrule
\ourscolor
\ourmethod{ClawBench} \textit{\small(Ours)}
  & \best{Real Web}
  & 153
  & \best{144}
  & \best{Write-heavy}
  & \best{Agent-as-Judge}
  & \best{5-Layer}
  & \best{\cmark} {\scriptsize(all tasks)} \\
\bottomrule
\end{tabular}
}
\vspace{2pt}

\end{table*}
\endgroup

We evaluate 8 frontier models under a shared browser-agent harness and find that current agents remain far from reliable real-world online assistants.
All models are instantiated in the same \textbf{OpenClaw browser-agent harness}, which provides a Chromium browser and fixed actions such as navigation, clicking, typing, scrolling, and page observation.
Claude Sonnet 4.6 reaches the highest success rate at only 33.3\%, followed by Qwen 3.5 at 26.1\% and GLM-5 at 24.2\%, while GPT-5.4 reaches only 6.5\%.
These scores are far below the same agents' reported results on established web-agent benchmarks such as OSWorld~\citep{xie2024osworld} and WebArena~\citep{zhou2024webarena}.

The trace analysis reveals that failures are not merely high-level reasoning errors.
Agents are frequently blocked by anti-bot and verification systems, produce non-human interaction patterns such as unnatural typing and missing mouse dynamics, refuse benign but consequential user tasks due to safety policies, or stop before the final transactional action.
These findings suggest that progress toward reliable online assistants requires not only better planning and perception, but also behaviorally grounded interaction models and robust handling of production-site constraints.

Our contributions are summarized as follows:

\noindent\textbf{(1)} We introduce \textsc{ClawBench}, a 153-task benchmark spanning 15 life categories and 144 live platforms, targeting write-heavy, state-changing workflows largely avoided by existing benchmarks.

\noindent\textbf{(2)} We design a safe live-web evaluation infrastructure based on final-request interception, allowing agents to interact with production websites while blocking the terminal HTTP request that would create real-world side effects.

\noindent\textbf{(3)} We develop a five-layer trajectory recording pipeline and an Agent-as-Judge evaluation protocol that compares agent runs against human reference trajectories and produces traceable binary verdicts.

\noindent\textbf{(4)} We benchmark eight frontier models under a shared browser-agent harness, showing that strong performance on existing web-agent benchmarks does not transfer to everyday write-heavy tasks on live production websites.

\noindent\textbf{(5)} We provide trace-level failure analysis revealing concrete, recurring bottlenecks in current web agents, including anti-bot blocking, unnatural interaction behavior, safety refusals, and incomplete transactional workflows.

\section{Benchmark}
\label{sec:benchmark}

\textsc{ClawBench} is designed to evaluate whether AI agents can complete \textbf{everyday, write-heavy online tasks} on \textbf{live production websites}.
It is built around three design choices: tasks are sourced from daily-life web use rather than synthetic page templates; agents interact with the live web while \textbf{final-request interception} blocks irreversible side effects; and each run is evaluated from \textbf{five-layer trace recording} by an \textbf{Agent-as-Judge}.
\autoref{tab:benchmark_comparison} positions \textsc{ClawBench} relative to prior web-agent benchmarks, and the overall pipeline (task definition, live-web execution, trace-based evaluation) is described in this section.

\paragraph{Positioning.}
\textsc{ClawBench} occupies a different point in the benchmark design space from static-trace datasets such as Mind2Web~\citep{deng2023mind2web}, self-hosted browser environments such as WebArena~\citep{zhou2024webarena} and VisualWebArena~\citep{koh2024visualwebarena}, and live-web benchmarks that primarily target read-only information seeking such as WebVoyager~\citep{he2024webvoyager} and Online-Mind2Web~\citep{xue2025illusion}.
Those settings are valuable for controlled evaluation, but they avoid many production-site constraints that matter for delegated online assistance.
\textsc{ClawBench} instead evaluates whether agents can carry user-specific information through live, write-heavy workflows while a task-scoped interceptor blocks the final irreversible request.

\subsection{Task Design and Collection}
\label{sec:task_design}

\textsc{ClawBench} focuses on tasks that ordinary users delegate in daily life but that existing web benchmarks often avoid: \textbf{write-heavy tasks} that submit information, reserve services, place orders, fill applications, create listings, or otherwise move a live website toward a state-changing final action.
These tasks are harder to evaluate than read-only search or sandbox clicking because success depends not only on finding the right page, but also on entering correct user-specific information, satisfying platform-specific validation, and reaching a terminal state that would normally modify server-side state.
This makes them a direct test of everyday online assistance rather than a proxy for web navigation alone.

We construct tasks from user-grounded website use and then convert them into auditable task cards.
Participants list websites they regularly use and propose realistic tasks involving information submission; annotators then formalize each candidate with a natural-language instruction, starting URL, required user files or profile fields, expected final state, and the terminal submission target used for safe evaluation.
The candidate pool is de-duplicated and manually filtered to remove unsuitable workflows, including read-only lookups, unavailable or geographically restricted sites, paid subscriptions, tasks requiring real monetary payment without a safe interception point, hard human-verification such as mandatory phone OTP or government-ID checks, and workflows whose irreversible effects cannot be controlled through a single terminal HTTP request.
For each retained task, a human annotator completes the workflow inside the same recording environment, producing a \textbf{human reference trajectory} and the corresponding terminal submission evidence.

\textsc{ClawBench} contains \textbf{153 tasks across 144 live production websites}, organized into 8 high-level category groups and 15 fine-grained categories, as shown in \autoref{fig:taxonomy}.
\label{sec:taxonomy}
This breadth-first design deliberately favors coverage across heterogeneous everyday platforms over repeated tasks on a small number of sites.
It tests whether agents can generalize across changing interfaces, authentication flows, form structures, validation logic, and anti-automation defenses, rather than overfitting to a few familiar benchmark environments.

\subsection{Safe Live-Web Execution}
\label{sec:interception}

The key safety challenge is that write-heavy tasks normally create \textbf{irreversible side effects} on production servers.
\textsc{ClawBench} addresses this with \textbf{final-request interception}: during the human reference run, an annotator identifies the \textbf{terminal HTTP request} that would commit the task, including its URL pattern, HTTP method, and required payload fields.
At runtime, a Chrome extension and CDP instrumentation server monitor outgoing HTTP requests; when an agent triggers a request matching the terminal submission target, the system captures the payload, blocks the request before it reaches the server, and records the interception event for evaluation.
All other traffic, including page loads, images, and analytics requests, proceeds normally, so the agent still faces the real interface dynamics of the live production website.

This design creates a \textbf{task-scoped safety envelope}, not a general guarantee that arbitrary browsing is side-effect-free.
Safety holds for the audited terminal endpoints under each task's annotated scope: the benchmark blocks the request that would submit the order, reservation, application, message, or other irreversible final action, while preserving the rest of the website interaction.
Accordingly, \textsc{ClawBench} excludes workflows whose side effects cannot be bounded by a single annotated terminal request, including uncontrolled account-setting changes, real payments or subscriptions without a safe interception point, hard human-verification gates, and chained irreversible intermediate writes.
In a validation study over all 153 tasks, the extension blocked the annotated terminal request in \textbf{100\%} of human reference runs with zero false positives on navigation traffic.

\subsection{Trace-Based Evaluation}
\label{sec:recording}
\label{sec:eval_protocol}

\textsc{ClawBench} records every run as synchronized evidence for both evaluation and diagnosis.
The \textbf{five-layer trace recording} consists of session video, action screenshots, HTTP traffic, agent messages, and browser actions: these layers capture what the agent saw, what it reasoned, what it did, and what network effects its actions produced.
Human annotators produce reference trajectories under the same recording setup, making agent and human runs comparable by page state, action sequence, submitted fields, and terminal payload.

\begin{figure}[!t]
\centering
\includegraphics[width=\columnwidth]{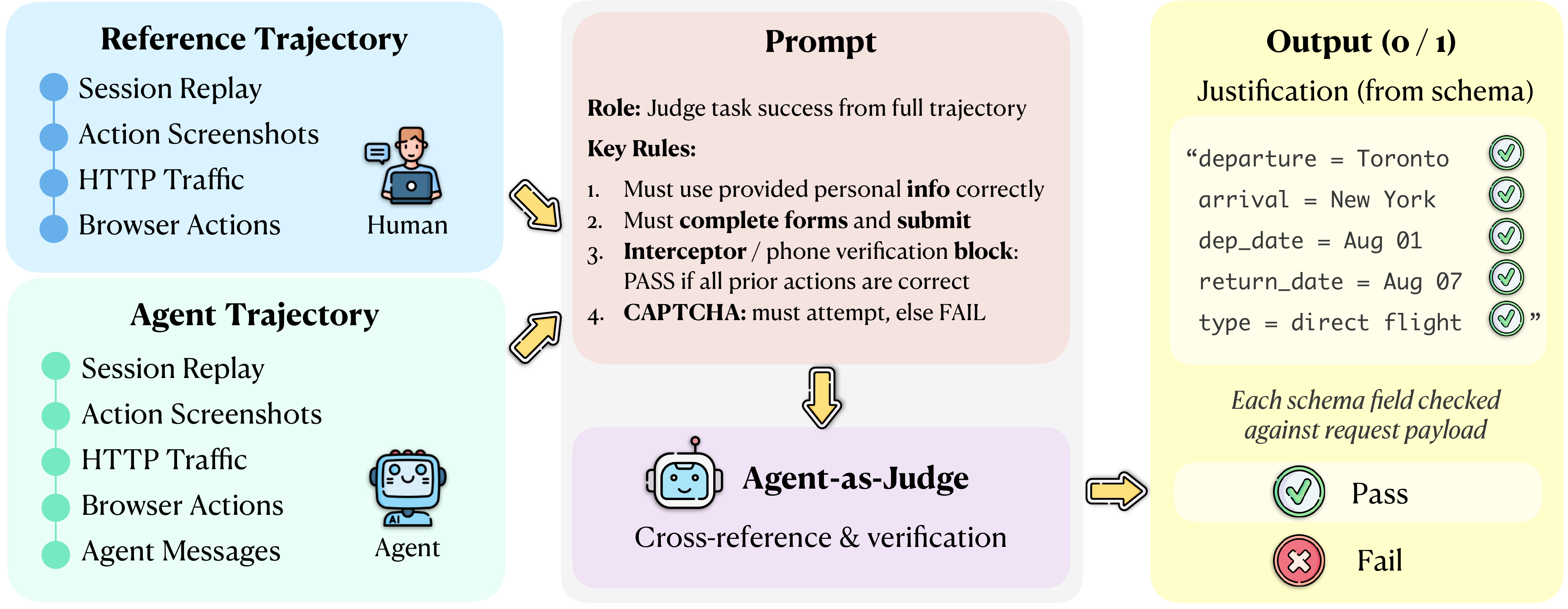}
\caption{
Agent-as-Judge inference pipeline. The evaluator compares the agent trace against the human reference trajectory across five evidence layers and returns a binary pass/fail verdict with a structured justification.
}
\label{fig:agentic_evaluator_inference}
\end{figure}

As shown in \autoref{fig:agentic_evaluator_inference}, each run is scored by Agent-as-Judge. In our implementation, Agent-as-Judge is instantiated with Claude Sonnet 4.6 as the underlying model and uses a fixed evaluation prompt for all agent runs.
It receives the task instruction, task-card constraints, the human reference trajectory, the agent trace, and the intercepted payload when available.
It returns a \textbf{binary pass/fail verdict} with a structured justification grounded in the recorded evidence.
The protocol is \textbf{outcome-oriented}: human reference trajectory specifies target final state and concrete field bindings, but the agent need not follow the same path if it reaches an equivalent terminal state.

For each task \(t \in T\), let \(q^{(t)}\) denote the task instruction, \(\mathcal{T}_a^{(t)}\) the recorded agent trajectory, and \(\mathcal{T}_h^{(t)}\) the recorded human reference trajectory.
Agent-as-Judge \(\mathcal{A}\) maps these inputs to a binary task-level verdict:
\begin{equation}
    \text{Score}(t) = \mathcal{A}\!\left(q^{(t)},\; \mathcal{T}_a^{(t)},\; \mathcal{T}_h^{(t)}\right),
\label{eq:scoring}
\end{equation}
where \(\text{Score}(t) \in \{0,1\}\).
The overall \textbf{success rate} over a task set \(T\) is
\begin{equation}
    \text{SR} = \frac{1}{|T|}\sum_{t \in T} \text{Score}(t).
\label{eq:success_rate}
\end{equation}
On human-adjudicated subsets covering two contrastive models
(\autoref{tab:judge_human_agreement}), Agent-as-Judge reaches
$84.97$--$93.46\%$ raw agreement with human verdicts.
Details of the prompt calibration procedure are provided in
Appendix~\ref{app:judge_calibration}.

\begin{table}[!htbp]
\caption{Agent-as-Judge vs.\ human judge on \textsc{ClawBench}. Agent SR and Human SR are pass rates from the Agent-as-Judge and human reviewer respectively over the same $153$ tasks; Agreement is the raw fraction of matching verdicts.}
\label{tab:judge_human_agreement}
\centering
\tablefontsize
\renewcommand{\arraystretch}{1.2}
\begin{tabular}{l r r r}
\toprule
\textbf{Model} & \textbf{Agent SR} & \textbf{Human SR} & \textbf{Agreement} \\
\midrule
Sonnet 4.6 & $33.33\%$ & $34.64\%$ & $93.46\%$ \\
GPT-5.4    & $6.54\%$  & $8.50\%$  & $84.97\%$ \\
\bottomrule
\end{tabular}
\end{table}

\section{Experiments}
\label{sec:experiments}

We evaluate current AI agents on the 153-task \textsc{ClawBench} benchmark.
This section first defines the shared experimental setup, then reports the main Agent-as-Judge success rates before turning to trace-level diagnosis in the next subsection.

\subsection{Experimental Setup}
\label{sec:experimental_setup}

\paragraph{Models.}
We evaluate 8 frontier models: Claude Sonnet 4.6~\citep{claude46sonnet}, GPT-5.4~\citep{gpt54}, Gemini 3.1 Flash Lite~\citep{gemini31flashlite}, Claude Haiku 4.5~\citep{claude45haiku}, Gemini 3 Flash~\citep{gemini3flash}, Gemini 3.1 Pro~\citep{gemini31pro}, GLM-5~\citep{zeng2026glm}, and Qwen 3.5~\citep{qwen35blog}.
All headline numbers in this section use this 8-model panel; additional released traces are used only for diagnostic analyses in the appendix.

\paragraph{OpenClaw harness.}
Each model is evaluated through the same OpenClaw harness controlling a Chromium browser in an isolated container.
The agent can interact with the browser through standard browser tools such as navigation, clicking, typing, scrolling, and page observation, while the \textsc{ClawBench} Chrome extension and CDP instrumentation record browser actions, screenshots, HTTP traffic, agent messages, and final-request interception data.
Chrome is launched with fixed flags and fresh profiles to reduce environmental variance across runs.

\paragraph{Metrics.}
Our primary metric is \textbf{success rate}: the percentage of tasks receiving a binary pass verdict from the Agent-as-Judge protocol described in \autoref{sec:eval_protocol}.
The judge compares each agent trajectory against the human reference trajectory and the intercepted final request when available.
We report success rate overall and across the eight high-level task category groups.

\subsection{Main Results}
\label{sec:main_results}

\begingroup
  \begin{table*}[!t]
  \centering
  \caption{Main results on \textsc{ClawBench}. Success rate (\%) of 8 models on \textsc{ClawBench}, reported overall and for each of the 8 high-level task categories. Avg. Cost denotes the average API cost per task, and Avg. Token denotes the average token consumption per task. \textbf{Bold} marks the best result; \underline{underline} marks the second best.}
  \label{tab:main_results}
  \vspace{4pt}
  \renewcommand{\arraystretch}{1.4}
  \newcommand{\osmodelicon}[2][1.05em]{%
    \IfFileExists{figures/icons/#2}{%
      \raisebox{-0.18\height}{\includegraphics[height=#1]{figures/icons/#2}}\,%
    }{}%
  }
    \adjustbox{max width=\textwidth}{
    \begin{tabular}{c l | w{c}{3.6em} | w{c}{3.8em} | w{c}{4.0em} | w{c}{3.2em} w{c}{3.2em} w{c}{3.2em} w{c}{3.2em} w{c}{3.6em} w{c}{3.2em} w{c}{3.2em} w{c}{4.2em}}
    \toprule
    \multirow{2}{*}{\textbf{Rank}} & \multirow{2}{*}{\textbf{Model}} & \multirow{2}{*}{\textbf{Overall}} & \multirow{2}{*}{\textbf{Avg. Cost}} & \multirow{2}{*}{\textbf{Avg. Token}} & \multicolumn{8}{c}{\textbf{Task Categories}} \\
    \cmidrule(lr){6-13}
     & & & & & \textbf{Daily} & \textbf{Finance} & \textbf{Work} & \textbf{Dev} & \textbf{Academic} & \textbf{Travel} & \textbf{Social} & \textbf{Pets} \\
  \altcolor
  1  & \osmodelicon{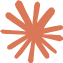}\,Claude Sonnet 4.6      & \textbf{33.3} & \$7.51 & 5.47M & \textbf{44.2} & \textbf{50.0} & 19.0 & 11.1 & \textbf{50.0} & \underline{23.1}  & \textbf{38.9} & \textbf{18.2} \\
  2  & \osmodelicon{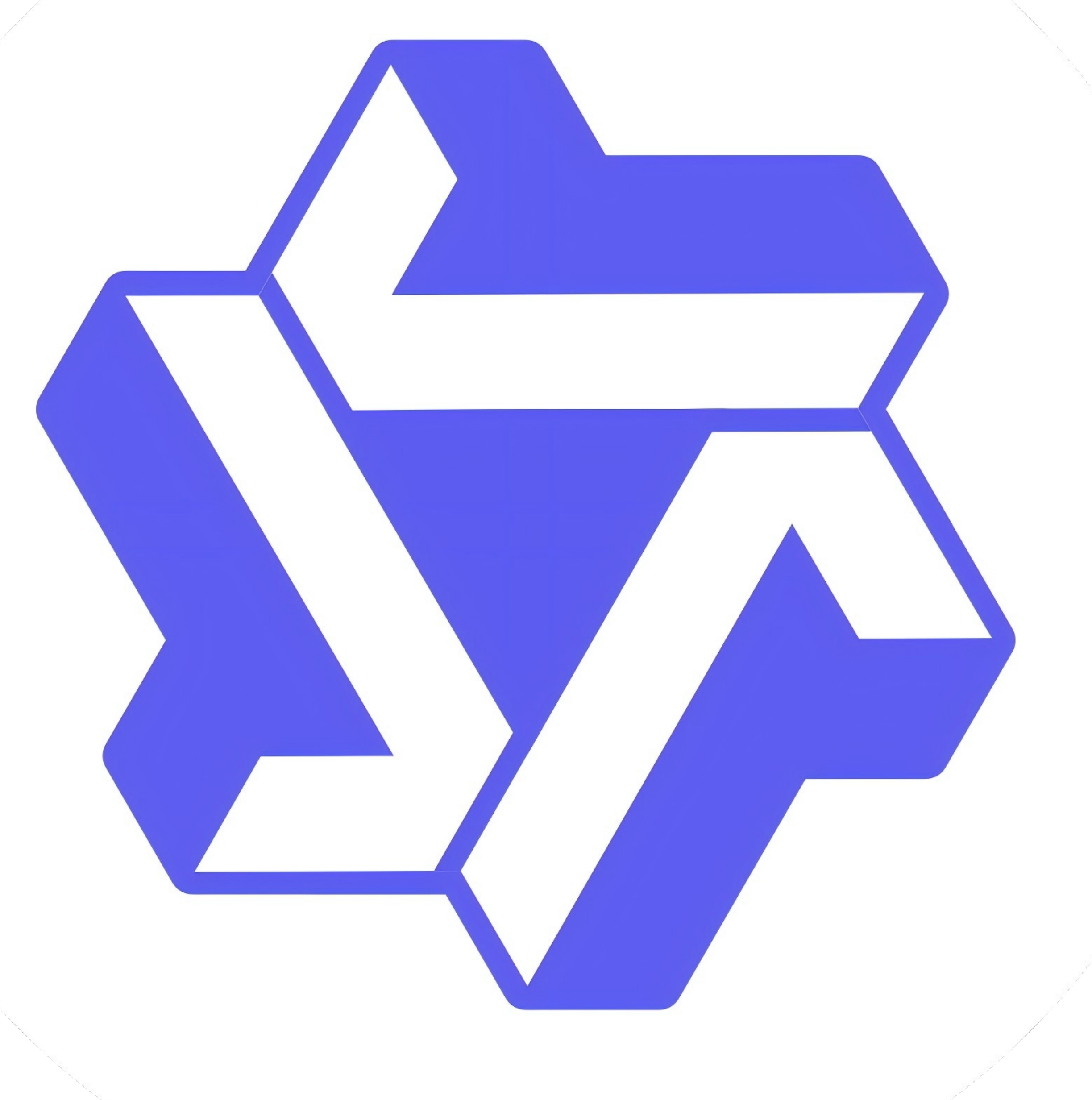}\,Qwen 3.5              & \underline{26.1} & \$1.02 & 2.52M & 25.0 & \textbf{50.0} & \underline{28.6} & \underline{22.2} & \underline{28.6} & \textbf{30.8} & \underline{22.2} & \textbf{18.2} \\
  \altcolor
  3  & \osmodelicon{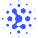}\,GLM-5       & 24.2 & \$0.64 & 3.33M & \underline{30.8} & 16.7 & \textbf{38.1} & 16.7 & \underline{28.6} & 0.0 & 16.7 & \textbf{18.2} \\
  4  & \osmodelicon{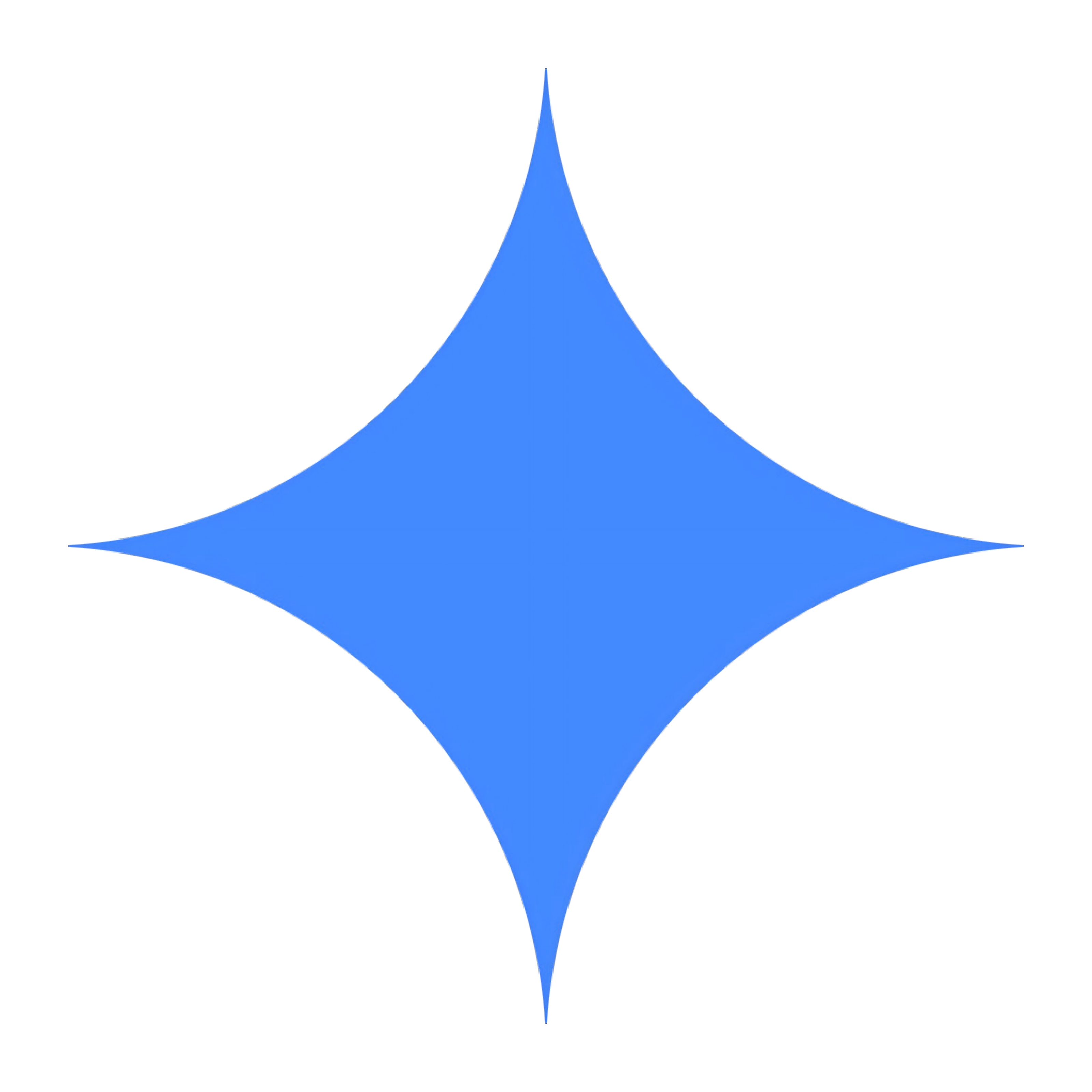}\,Gemini 3 Flash        & 19.0 & \$1.55 & 5.81M & 15.4 & \underline{33.3}  & 23.8 & \underline{22.2} & \underline{28.6} & \textbf{30.8} & 11.1 & 0.0 \\
  \altcolor
  5  & \osmodelicon{claude-logo.pdf}\,Claude Haiku 4.5      & 18.3 & \$1.68 & 4.56M & 15.4 & \underline{33.3}  & 19.0 & \textbf{27.8} & 21.4 & 7.7 & 16.7 & \textbf{18.2} \\
  6  & \osmodelicon{gemini.pdf}\,Gemini 3.1 Pro        & 9.8 & \$3.34 & 3.16M & 5.8 & 16.7 & 9.5 & \underline{22.2} & 14.3 & 15.4 & 5.6 & 0.0 \\
  \altcolor
  7  & \osmodelicon{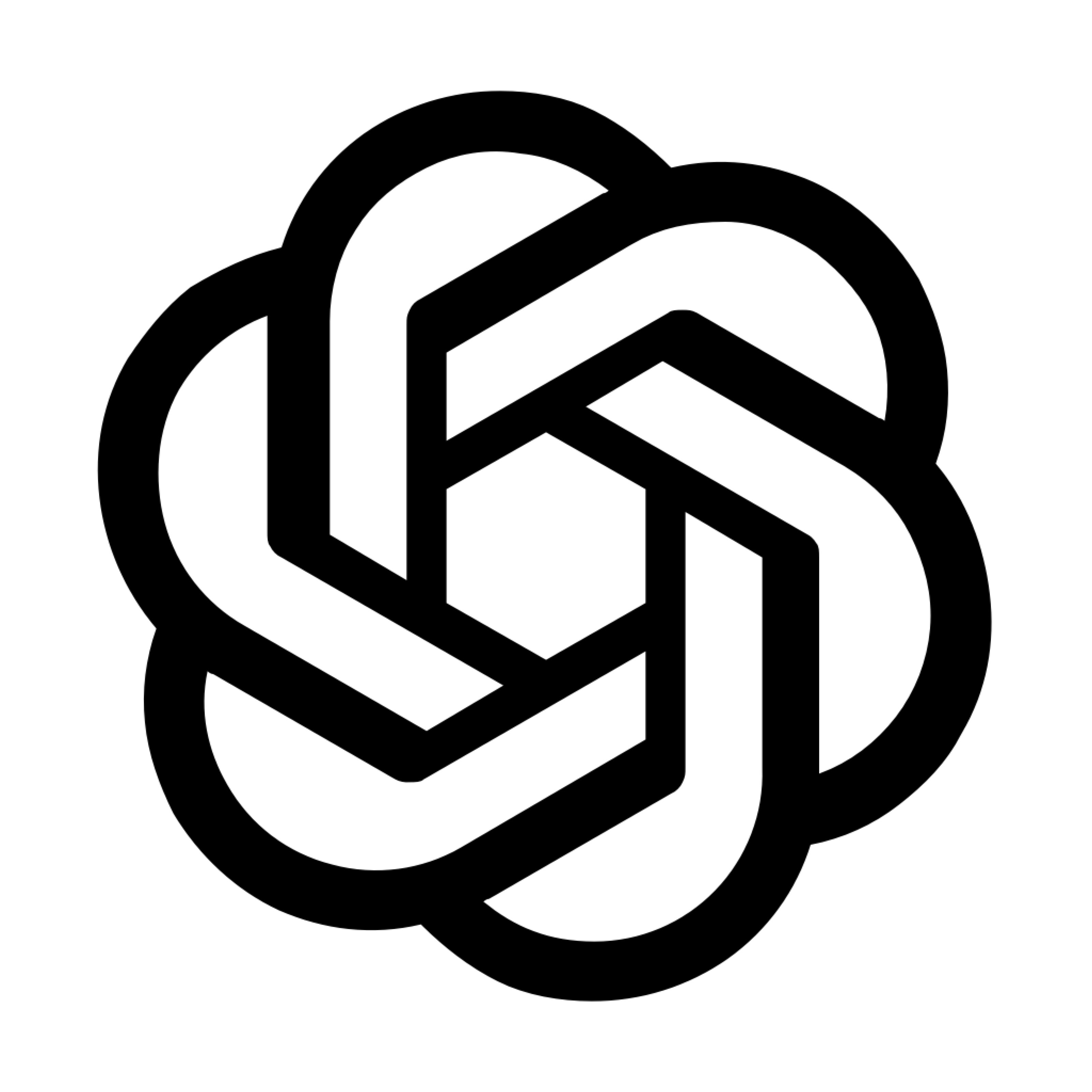}\,GPT-5.4               & 6.5 & \$1.91 & 2.66M & 9.6 & 0.0 & 0.0 & 11.1 & 7.1 & 7.7 & 0.0 & \underline{9.1} \\
  8  & \osmodelicon{gemini.pdf}\,Gemini 3.1 Flash Lite & 3.3 & \$0.11 & 1.18M & 1.9 & 0.0 & 0.0 & 5.6 & 14.3 & 0.0 & 0.0 & \underline{9.1} \\
  \bottomrule
  \end{tabular}
  }
  \end{table*}
\endgroup

\paragraph{Overall success rate.}
\autoref{tab:main_results} shows that \textsc{ClawBench} remains far from saturated.
Claude Sonnet 4.6 is the strongest model, but its overall success rate is only 33.3\%; the next best model, Qwen 3.5, reaches 26.1\%, followed by GLM-5 at 24.2\%.
The large drop from frontier-agent performance on established web-agent benchmarks to these live, write-heavy tasks suggests that everyday online task completion remains a substantially harder setting than controlled browser evaluation.

\paragraph{Task-level saturation.}
The low aggregate scores are not driven by a small set of outlier tasks.
Across the 8-model panel, 68 of 153 tasks (44.4\%) are solved by no model, only one task is solved by 7 of 8 models, and no task is solved by all eight.
This shows that \textsc{ClawBench} is far from saturated at the task level, not only at the leaderboard level.

\paragraph{Model ranking and domain variation.}
The leaderboard is not explained by a single global capability.
Sonnet 4.6 leads the overall leaderboard and is strongest on Daily, Academic, and Social tasks, while tying Qwen 3.5 on Finance and Pets.
Qwen 3.5 is the second-best overall model and ties Gemini 3 Flash on Travel, GLM-5 is strongest on Work, and Claude Haiku 4.5 is strongest on Dev.
This uneven profile indicates that current agents do not yet transfer uniformly across everyday domains: the same model that handles shopping or social workflows may still struggle with job applications, travel booking, or developer workflows.
Appendix~\ref{app:category-breakdowns} provides the fine-grained category table and rank-by-category visualization.

\begin{figure*}[!htbp]
\centering
\includegraphics[width=\linewidth]{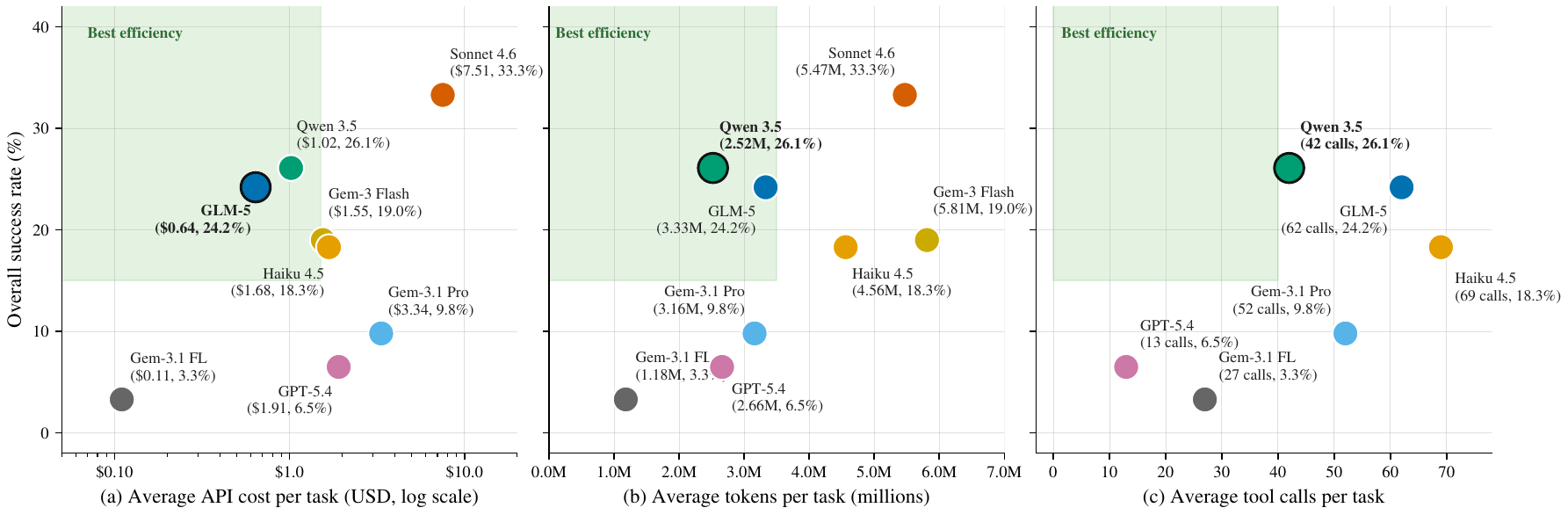}
\caption{Operating-cost efficiency on \textsc{ClawBench}. Overall success rate plotted against three operating-cost axes: \textbf{(a)} average API cost per task in USD (log scale),
\textbf{(b)} average tokens consumed per task in millions, and \textbf{(c)} average tool calls per task. The shaded region in each panel marks an efficient operating regime (SR $\ge 15\%$ at low cost on that axis). The bold-outlined dot highlights the most efficient capable model per panel.
}
\label{fig:cost_pareto}
\end{figure*}

\paragraph{Cost-performance tradeoff.}
GLM-5 is the most cost-efficient on \textsc{ClawBench} at $\$0.64$ per task and $24.2\%$ SR; Claude Sonnet 4.6 holds the accuracy ceiling at $33.3\%$ SR but pays $12\times$ that cost for $9.1$\,pp more SR (\autoref{tab:main_results}, \autoref{fig:cost_pareto}). Token usage tells a similar story but with a different winner: Qwen 3.5 reaches $26.1\%$ SR on just $2.52$M tokens per task, the lowest per-SR-point token budget in the panel, while Gemini 3 Flash spends $5.81$M tokens for only $19.0\%$ SR. 
Per-trajectory tool-call counts (\autoref{fig:cost_pareto} panel c) further differentiate the panel: Qwen 3.5 is the most action-efficient capable model at $42$ average tool calls per task for $26.1\%$ SR, while GPT-5.4 averages only $13$ tool calls per task at $6.5\%$ SR, reflecting early termination rather than efficient task completion. Beyond Qwen 3.5 and the two early-terminating low-SR models, the other higher-interaction agents cluster around 52 to 86 average tool calls per task. This non-monotonic frontier shows that higher spending alone does not reliably translate into better completion on live, write-heavy web tasks. Taken together, the tradeoff suggests two clear operating points: GLM-5 or Qwen 3.5 for budget-constrained evaluation, and Sonnet 4.6 when maximizing accuracy is the primary goal.

\subsection{Trace-Level Diagnosis}
\label{sec:analysis}
\label{sec:trace_diagnosis}

The headline success rates show that current agents struggle, but the trace corpus explains what kind of struggle it is.
This subsection keeps the diagnosis focused on four non-overlapping signals: how much interaction failing trajectories consume, what kinds of browser actions those trajectories contain, how far agents progress through write-heavy workflows, and how agent interaction dynamics differ from human behavior.
Additional pass/fail distribution plots, stop-reason diagnostics, and failure-case evidence are reported in Appendix~\ref{sec:appendix-extended}.

\begin{figure*}[t]
\centering
\includegraphics[width=0.98\linewidth]{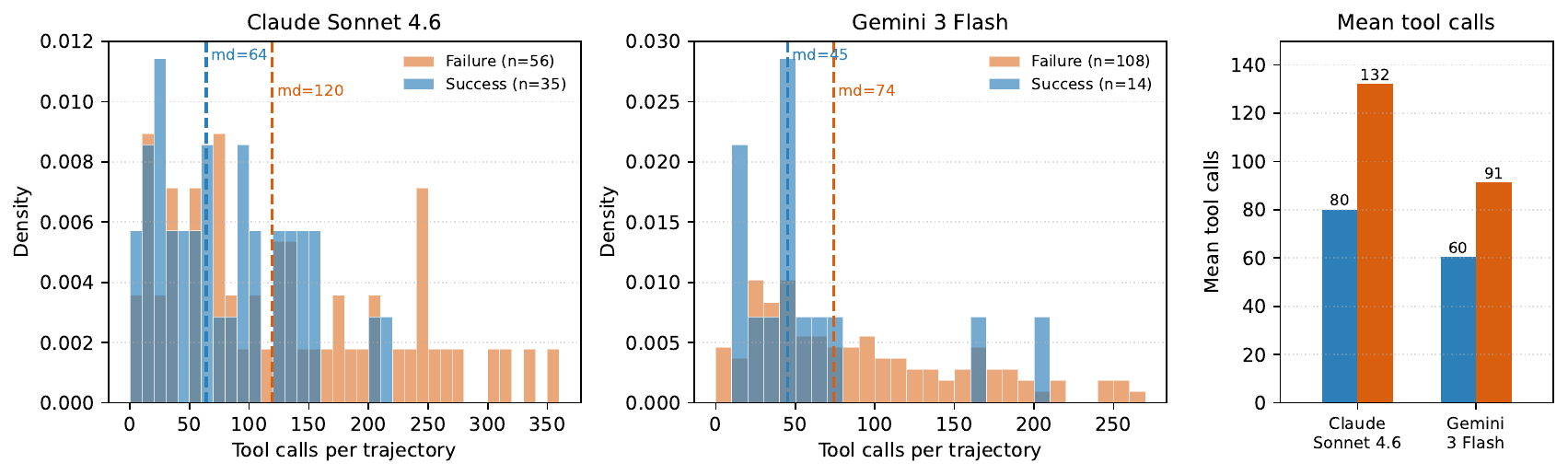}
\caption{Tool-call counts by outcome for Claude Sonnet 4.6 and Gemini 3 Flash. Dashed lines mark medians; failed trajectories make more calls and have longer tails, indicating unproductive rather than insufficient exploration.}
\label{fig:toolcalls}
\end{figure*}

\paragraph{Failed trajectories often consume more browser interaction.}
We count browser tool calls from the agent message log, excluding passive page reloads that would inflate the raw action stream.
\autoref{fig:toolcalls} compares a strong model, Claude Sonnet 4.6, with a mid-tier model, Gemini 3 Flash, on trajectories that actually started.
For both models, failed trajectories use more tool calls than successful ones: Sonnet 4.6 has a median of 64 tool calls on successful runs and 120 on failed runs, while Gemini 3 Flash has a median of 45 on successful runs and 74 on failed runs.
Failures also carry a heavy right tail, reaching 354 tool calls for Sonnet 4.6 and 263 for Gemini 3 Flash.
This pattern rules out a simple ``not enough exploration'' explanation: many failures are unproductive retry loops on anti-bot walls, repeated re-snapshotting, or repeated attempts to reach a final submission state.

\begin{figure*}[!tbp]
\centering
\begin{subfigure}[t]{0.49\textwidth}
    \centering
    \includegraphics[width=\linewidth,height=0.19\textheight,keepaspectratio]{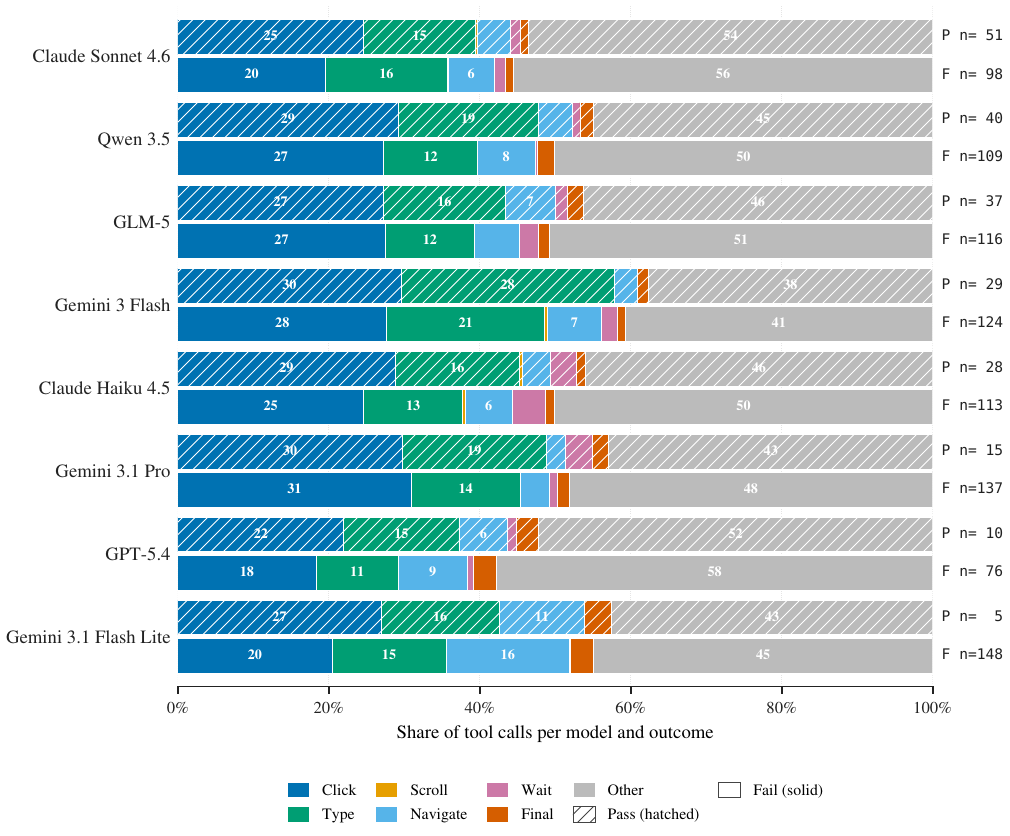}
    \caption{Action composition by outcome.}
    \label{fig:tool_composition}
\end{subfigure}
\hfill
\begin{subfigure}[t]{0.49\textwidth}
    \centering
    \includegraphics[width=\linewidth,height=0.19\textheight,keepaspectratio]{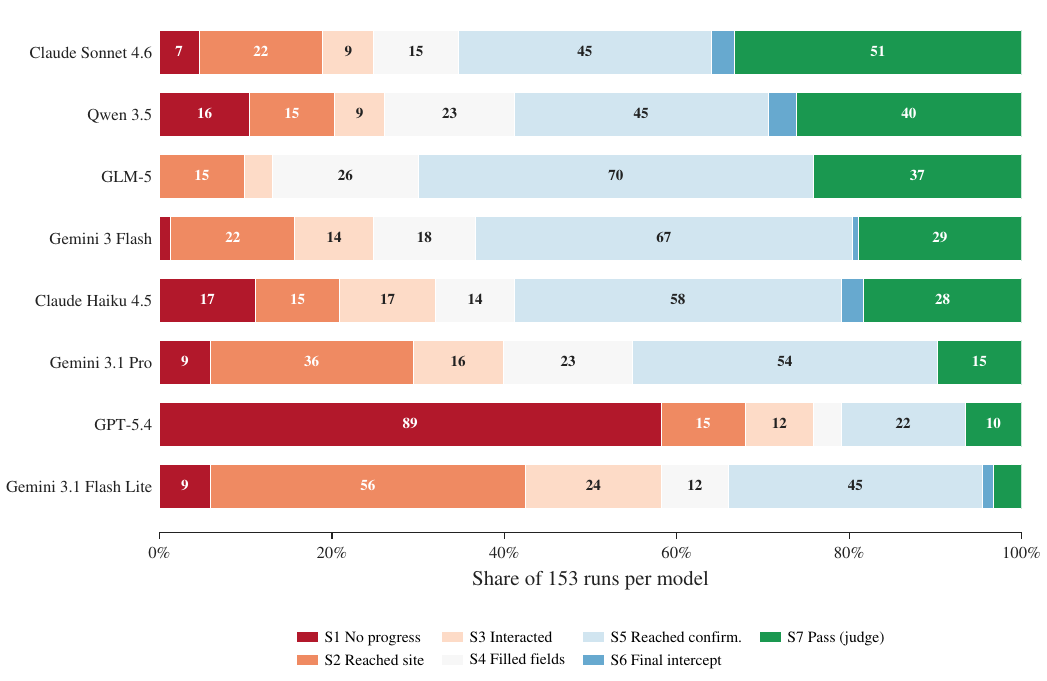}
    \caption{Deepest workflow stage reached.}
    \label{fig:appendix-stage-funnel}
\end{subfigure}
\caption{Two views of last-mile hesitation. (a) Tool-use composition by model and outcome; runs with zero tool calls are excluded from this view and are analyzed separately. (b) Deepest workflow stage over the 1{,}224 \textsc{ClawBench} runs. Failures often interact substantially yet stop before the terminal step.}
\end{figure*}

\paragraph{Tool-use composition and stage position reveal last-mile hesitation.}
Tool-call volume alone does not say whether agents are navigating, filling fields, or attempting the final state-changing action.
\autoref{fig:tool_composition} therefore decomposes every \texttt{toolCall} in agent messages into click, type, scroll, navigate, wait, final, and other categories, grouped by model and outcome.
\autoref{fig:appendix-stage-funnel} complements this action-type view by assigning each of the 1{,}224 runs to its deepest observed workflow stage.
The main pattern is not a large shift in ordinary click/type behavior: Sonnet 4.6 issues 25\% / 15\% click/type on passes and 20\% / 16\% on failures, while Gemini 3 Flash issues 30\% / 28\% on passes and 28\% / 21\% on failures.
Instead, failing runs combine higher interaction volume with a smaller terminal-commitment slice: the final click or type action is consistently less represented on failures than on passes.
The stage funnel localizes the same effect procedurally: six of eight models concentrate many failures at S5, where the agent reaches the final confirmation step but does not commit the state-changing request.
Together, these views connect the effort signal above to the write-heavy nature of \textsc{ClawBench}: agents often interact with the page but hesitate before the final Submit, Confirm, or Place-Order step.

\begin{figure}[!tbp]
\centering
\includegraphics[width=\linewidth]{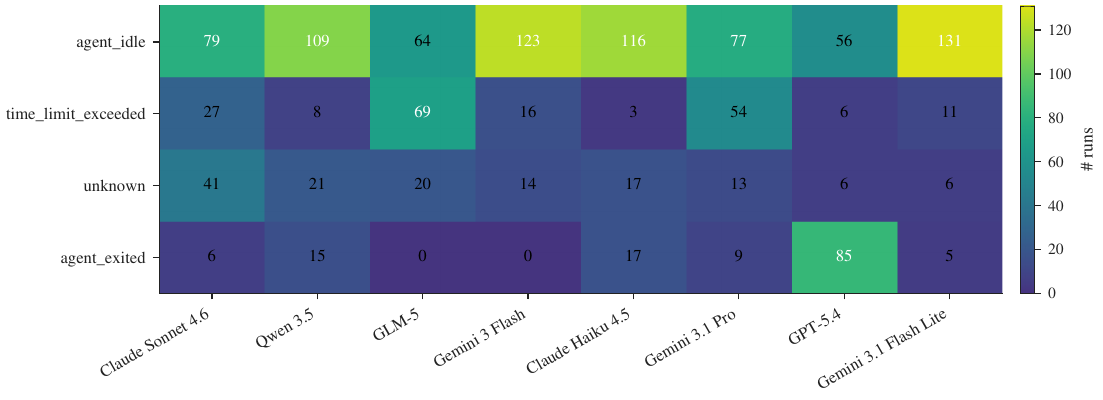}
\caption{\textbf{Termination reason by model.} Counts of trajectory terminations across \texttt{agent\_idle}, \texttt{time\_limit\_exceeded}, \texttt{agent\_exited}, and \texttt{unknown} for the eight frontier agents.}
\label{fig:stop_reason}
\end{figure}

\paragraph{Failure regimes differ across agents.}
The same low success rate can reflect different operational failures.
Gemini 3.1 Flash Lite terminates with \texttt{agent\_idle} on 131 of 153 runs, GLM-5 reaches the 30-minute wall-clock limit on 69 runs, and GPT-5.4 exits through \texttt{agent\_exited} on 85 runs (\autoref{fig:stop_reason}).
Thus progress on live-web agents should be measured along separate axes: preventing premature exit, sustaining action through live-site friction, and committing safely at the final step.

\section{Conclusion}
\label{sec:conclusion}
We introduce \textsc{ClawBench}, a benchmark of 153 real-world everyday web tasks spanning 144 live platforms across 8 high-level category groups. By evaluating agents on live production websites and focusing on write-heavy, state-changing workflows, \textsc{ClawBench} provides a more realistic testbed than prior benchmarks built on static pages or sandboxes. Our framework combines final-request interception, 5-layer trajectory recording, and Agent-as-Judge evaluation. Experiments on 8 models show that strong performance on existing web-agent benchmarks does not transfer to \textsc{ClawBench}, highlighting the gap between controlled benchmark success and real-world web competence. We release the benchmark, evaluation toolkit, infrastructure, and agent trajectories to support future research on realistic web-agent evaluation. Ultimately, we hope \textsc{ClawBench} helps guide the next generation of web agents toward reliably and safely completing the everyday online tasks people most want to delegate, turning today's promising demonstrations into dependable assistance in daily life.

\UseLocalInputs

\bibliography{colm2026_conference}
\bibliographystyle{colm2026_conference}

\newpage

\UseProjectInputs
\section*{Limitations}
\label{sec:limitations}

\textsc{ClawBench} evaluates agents on live production websites, trading
rerun determinism for ecological validity. Sites change layouts, run
A/B tests, vary by region or account state, and deploy anti-automation
defenses, so exact reproduction is not guaranteed; we mitigate this by
recording human reference trajectories, intercepted terminal payloads,
screenshots, browser actions, HTTP traffic, and agent messages, so that
individual outcomes remain auditable even when site state drifts. Each
run is correspondingly more expensive than static-trace or sandbox
benchmarks, it requires a live session, repeated model calls, network
instrumentation, and trace-based judging, which limits the number of
repeated trials, prompt variants, and ablations we report under a fixed
budget. Results are also conditioned on a shared OpenClaw harness,
prompt, browser configuration, and action interface: the scores measure
each model within this agent stack rather than the language model in
isolation, and a different scaffold may yield different outcomes on the
same tasks.

Our primary metric is a binary pass/fail verdict from an Agent-as-Judge
protocol grounded in human references and multi-layer traces. Binary
scoring is conservative for consequential write-heavy tasks, where
partial progress rarely suffices if the final submission is missing,
malformed, or sent to the wrong endpoint, but it abstracts away
intermediate behavior such as reaching the target site, filling most
fields correctly, or stopping at a verification gate; the released
traces and structured justifications support complementary analyses of
these behaviors. Because one evaluated model (Claude Sonnet 4.6) shares a model family with the judge, we audit per-model human agreement in Appendix~\ref{app:safety-audits}; Sonnet 4.6 reaches $93.46\%$ agreement and GPT-5.4 reaches $84.97\%$, and an expanded four-model audit covers two additional models. The suite is also broad but not exhaustive: 153 tasks
across 144 platforms emphasize cross-site generalization over dense
per-domain coverage, and underrepresent mobile-only, non-English, and
accessibility-dependent workflows, which future versions can extend
under the same task-card, interception, and trace-based protocol.

\appendix

\colorlet{clawpromptblue}{appleindigoIC}
\colorlet{clawpromptbg}{appleIndigoLight}

\newtcolorbox{clawpromptbox}[1]{
  enhanced,
  colback=clawpromptbg,
  colframe=clawpromptblue,
  colbacktitle=clawpromptblue,
  coltitle=white,
  fonttitle=\bfseries\footnotesize,
  fontupper=\footnotesize,
  title={#1},
  boxrule=0.7pt,
  arc=1pt,
  outer arc=1pt,
  left=4pt,
  right=4pt,
  top=4pt,
  bottom=4pt,
  lefttitle=4pt,
  righttitle=4pt,
  toptitle=2pt,
  bottomtitle=2pt,
  before skip=0.65em,
  after skip=0.8em
}

\section{Related Work}
\label{sec:related_work}

\paragraph{Web Agent Benchmarks.}
Early web agent benchmarks such as MiniWoB \citep{shi2017miniwob} evaluated agents on simplified, synthetic web interfaces with short action sequences.
WebArena \citep{zhou2024webarena} introduced self-hosted, realistic web environments with 812 tasks across 5 domains, using URL and element matching for evaluation.
VisualWebArena \citep{koh2024visualwebarena} extended this to visually grounded tasks on 3 self-hosted sites.
Mind2Web \citep{deng2023mind2web} scaled to 2{,}350 tasks on 137 real-world domains but evaluated action sequences rather than end-to-end task completion.
OSWorld \citep{xie2024osworld} broadened the scope to full operating system tasks across 9 applications.
More recently, REAL Bench \citep{garg2025real} evaluated agents on live websites but relied on manual rating for scoring.
\textsc{ClawBench} differs from all prior work by (i) operating on 144 live platforms rather than self-hosted sandboxes, (ii) focusing on write-heavy, state-changing tasks, and (iii) providing traceable, comparative evaluation against human reference trajectories through Agent-as-Judge.

\paragraph{LLM-Based Web Agents.}
The emergence of large language models has driven rapid progress in autonomous web agents.
Systems such as WebGPT \citep{nakano2021webgpt}, WebAgent \citep{gur2023webagent}, and SeeAct \citep{zheng2024seeact} demonstrated that LLMs can interpret web pages and execute multi-step browsing tasks when given appropriate observation and action interfaces.
Recent approaches combine visual perception (screenshots) with structured page representations (accessibility trees, HTML) to improve grounding accuracy.
BrowserAgent \citep{yu2025browseragent} proposes a scalable training framework for web agents that operate directly on raw web pages via human-inspired browser actions such as scrolling, clicking, and typing.
Agent frameworks including AgentGPT, AutoGPT, and OpenClaw provide standardized interfaces for deploying LLMs as web agents with tool use and action execution capabilities.
\textsc{ClawBench} is designed to evaluate any agent system that can control a Chromium browser, independent of the underlying model or framework.

\paragraph{Evaluation Methods for Agent Systems.}
Evaluating autonomous agents remains challenging due to the diversity of possible action trajectories and the difficulty of defining success criteria.
Prior work has used action sequence matching, URL-based success detection, screenshot comparison, and human judgement.
Action-level metrics suffer from the problem of multiple valid paths: an agent may complete a task correctly through a different sequence of actions than the reference trajectory.
Screenshot-based methods require visual similarity thresholds that introduce non-determinism.
Human evaluation, while flexible, is expensive and non-reproducible.
\textsc{ClawBench} sidesteps these issues by combining intercepted submission payloads with Agent-as-Judge, which performs explicit step-level alignment between the agent trajectory and a human reference trajectory, producing a binary verdict together with a structured justification grounded in the recorded evidence.

\paragraph{Concurrent and Complementary Work.}
Several recent benchmarks address related but distinct aspects of web agent evaluation.
TheAgentCompany \citep{xu2024theagentcompany} provides a self-hosted sandbox simulating a software company with 175 tasks and checkpoint-based partial credit; \textsc{ClawBench} trades environmental control for real-world breadth across 144 live platforms.
EconWebArena \citep{liu2025econwebarena} is a live-web benchmark for economic research tasks featuring 360 read-only tasks with exact numeric matching and URL provenance; \textsc{ClawBench} extends the live-web paradigm to write-heavy, state-changing tasks.
MCP-Bench \citep{wang2025mcp} evaluates LLM agents on tool invocation via the Model Context Protocol with strict schema validation; \textsc{ClawBench} targets browser-based web interaction rather than structured API calls.
SwingArena \citep{xu2025swingarena} introduces an innovative competitive programming arena that rigorously evaluates agents on long-context GitHub issue solving, advancing code-centric agent benchmarking; \textsc{ClawBench} extends this evaluation spirit to everyday web interactions on live platforms beyond the software engineering domain.
TrickyArena \citep{ersoy2025investigating} studies dark pattern susceptibility in web agents across 4 controlled applications---an orthogonal safety concern that highlights the importance of evaluating on real websites where dark patterns occur naturally.
AssistantBench \citep{yoran2024assistantbench} defines 214 realistic open-web tasks with automated evaluation, focusing on information retrieval; \textsc{ClawBench} complements this with write-heavy tasks.
WebCanvas \citep{pan2024webcanvas} proposes key-node evaluation for 542 tasks on dynamic websites in a similar live-web setting but without HTTP payload verification.
Taken together, these efforts illustrate a fundamental realism-vs-reproducibility trade-off: sandboxed benchmarks offer perfect reproducibility but may not reflect the complexity of real websites, while live-web benchmarks expose agents to authentic challenges at the cost of environmental variability. \textsc{ClawBench} deliberately chooses realism and mitigates reproducibility concerns through human-grounded comparative evaluation and full multi-layer trajectory recording.

\paragraph{Deep Research and Code Agent Benchmarks.}
Beyond web browsing, recent work has advanced evaluation for deep research agents and software engineering tasks.
Dr.~Bench \citep{yao2025drbench} introduces a multidimensional evaluation framework for deep research agents with 214 expert-curated tasks across 10 domains, targeting long-form report generation.
BrowseComp-Plus \citep{chen2025browsecompplus} provides a fair and transparent benchmark for deep research systems by employing a fixed, curated corpus with human-verified supporting documents, enabling disentangled analysis of retrieval and reasoning components.
In the software engineering domain, SWE-Next \citep{liang2026swenext} presents an execution-grounded framework for scalable SWE task and trajectory collection from real merged pull requests, while SWE-QA-Pro \citep{cai2026sweqapro} proposes a representative benchmark for repository-level code understanding constructed from diverse, long-tail repositories.
The recent Claw-Eval \citep{ye2026claweval} introduces trajectory-aware grading and safety/robustness evaluation for autonomous agents---representing an evolution toward trustworthy evaluation paradigms beyond earlier-generation benchmarks.
While these benchmarks focus on research synthesis or code-centric tasks, \textsc{ClawBench} targets a complementary and equally important domain: the everyday web tasks that ordinary users need to accomplish regularly in their lives and work.

\section{Extended Trace-Level Analysis}
\label{sec:appendix-extended}

The trace corpus extends the main-paper evaluation on the 1{,}224 main agent runs (8 models $\times$ 153 tasks) with per-run behavioral signals: \texttt{actions.jsonl}, \texttt{agent-messages.jsonl}, \texttt{interception.json}, judge rationales, and browser-level event logs.
The appendix uses these traces as supporting evidence for the main findings rather than as a second leaderboard.
All headline success rates remain the 8-model Agent-as-Judge results in \autoref{tab:main_results}.

\subsection{Safety and Evaluator Audits}
\label{app:safety-audits}

\paragraph{Safety-envelope audit.}
We audit the final-request interception envelope along three dimensions.
\emph{First}, on the 153 human reference runs, the harness intercepted every annotated terminal request that the reference reached, with zero false-positive blocks on the recorded navigation traffic; this confirms the interceptor's coverage at the trace level.
\emph{Second}, on the 8-model agent panel, 111 of 1{,}224 runs ($9.1\%$, spanning 35 distinct tasks) triggered final-request interception.
\emph{Third}, among 177{,}687 non-terminal \texttt{POST}/\texttt{PUT}/\texttt{PATCH}/\texttt{DELETE} requests, 57.1\% are third-party tracking and ad-tech, 41.5\% are CDN challenges and GraphQL-as-\texttt{POST} persisted-query reads, 0.7\% are login/auth, 0.2\% are cart/session UI state, and 0.2\% are form validation.
The remaining 0.26\% (464 requests) carry commit-suggestive tokens such as \texttt{submit}, \texttt{checkout}, or \texttt{payment}; manual inspection finds Shopify checkout proposals, Stripe/Braintree Elements bootstraps, and quote previews, none of which commit an irreversible task side effect.

\subsection{Task-Level and Category Breakdowns}
\label{app:category-breakdowns}
\label{app:fine-category}

\autoref{tab:category_fine} expands the eight high-level categories in \autoref{tab:main_results} into the 15 fine-grained categories used in the benchmark taxonomy.
This table is a diagnostic breakdown for the six models covered by the fine-category coding pass; the main-paper leaderboard remains the authoritative 8-model result.

\begin{table*}[tbp]
\centering
\tablefontsize
\caption{Category-wise success rate across 15 task categories.
Each value denotes the fraction of successfully completed tasks within the corresponding fine-grained category. \textbf{Bold} marks the best-performing model within each category.}
\label{tab:category_fine}
\vspace{4pt}
\renewcommand{\arraystretch}{1.35}
\adjustbox{max width=\textwidth}{
\begin{tabular}{l c c c c c c c}
\toprule
\shortstack[c]{Category} & \shortstack[c]{Tasks}
& \shortstack[c]{Claude\\Sonnet 4.6}
& \shortstack[c]{GLM-5}
& \shortstack[c]{Gemini\\3 Flash}
& \shortstack[c]{Claude\\Haiku 4.5}
& \shortstack[c]{GPT-5.4}
& \shortstack[c]{Gemini 3.1\\Flash Lite} \\
\midrule
\altcolor
Education & 9 & 0.44 & 0.33 & 0.33 & 0.33 & 0.11 & 0.11 \\
Personal Mgmt & 4 & 0.25 & \best{0.75} & 0.50 & 0.25 & 0.00 & 0.00 \\
\altcolor
Office & 9 & 0.33 & \best{0.56} & 0.22 & 0.22 & 0.00 & 0.00 \\
Shopping & 16 & \best{0.62} & 0.31 & 0.19 & 0.06 & 0.06 & 0.00 \\
\altcolor
Social & 8 & \best{0.75} & 0.12 & 0.12 & 0.25 & 0.00 & 0.00 \\
Finance & 6 & \best{0.50} & 0.17 & 0.33 & 0.33 & 0.00 & 0.00 \\
\altcolor
Entertainment & 15 & \best{0.33} & 0.27 & 0.13 & 0.13 & 0.07 & 0.00 \\
Daily Life & 21 & \best{0.33} & 0.14 & 0.10 & 0.14 & 0.10 & 0.05 \\
\altcolor
Academia & 5 & \best{0.60} & 0.20 & 0.20 & 0.00 & 0.00 & 0.20 \\
Dev \& Tech & 15 & 0.13 & 0.20 & 0.27 & \best{0.27} & 0.13 & 0.07 \\
\altcolor
Pets & 11 & \best{0.18} & 0.09 & 0.00 & \best{0.18} & 0.09 & 0.09 \\
Rating & 10 & 0.10 & \best{0.20} & 0.10 & 0.00 & 0.00 & 0.00 \\
\altcolor
Travel & 13 & 0.15 & 0.00 & 0.15 & 0.08 & 0.08 & 0.00 \\
Automation & 3 & 0.00 & 0.00 & \best{0.33} & \best{0.33} & 0.00 & 0.00 \\
\altcolor
Job Search & 8 & 0.00 & 0.00 & 0.00 & \best{0.12} & 0.00 & 0.00 \\
\bottomrule
\end{tabular}
}
\end{table*}

\begin{figure*}[tbp]
\centering
\includegraphics[width=\linewidth]{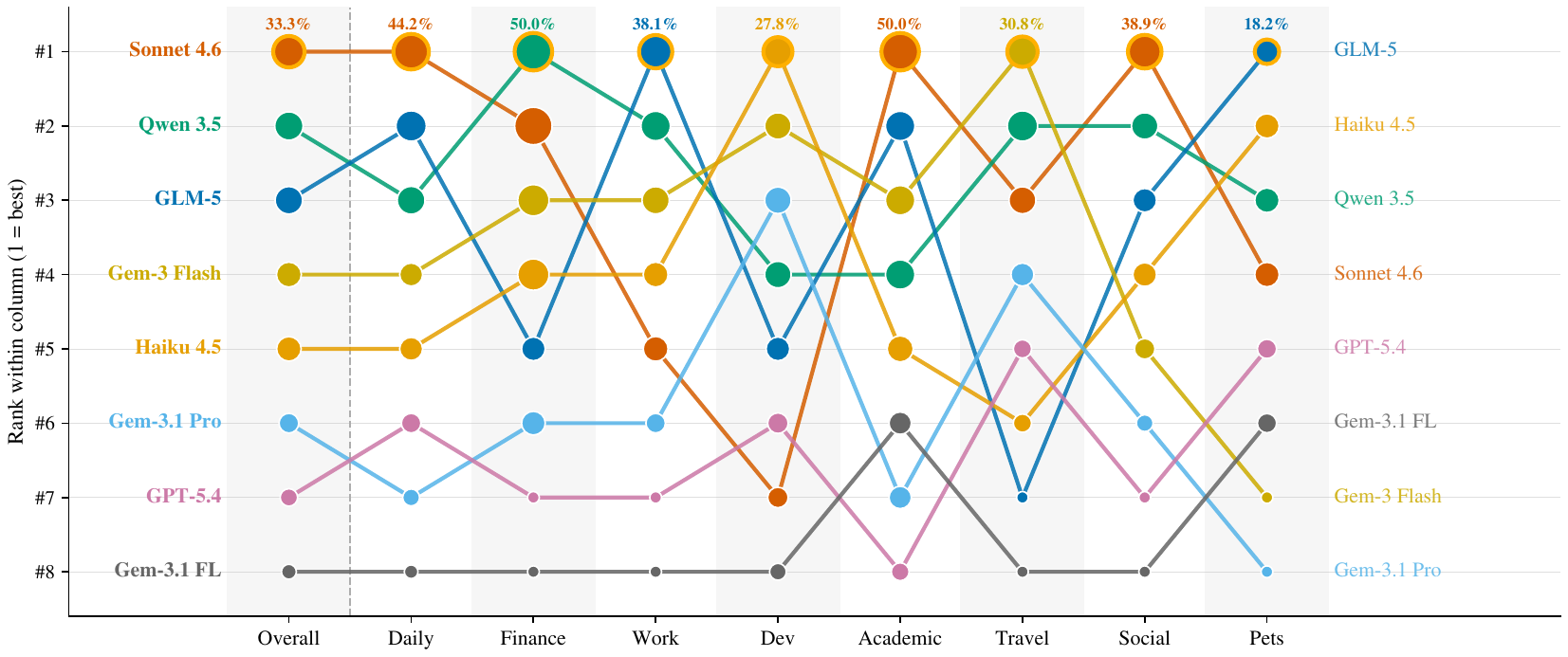}
\caption{\textbf{Category leadership rotates across the eight-model panel.} Per-column rank of each model on overall success rate and the eight high-level task categories. Marker size scales with success rate; gold-outlined markers mark the rank-1 model in each column.}
\label{fig:appendix-bump-chart}
\end{figure*}

\paragraph{Category leadership.}
\autoref{fig:appendix-bump-chart} re-plots \autoref{tab:main_results} as per-column ranks over overall success rate and the eight high-level categories.
It shows why the main text treats the leaderboard and domain variation separately: Sonnet 4.6 leads four of eight categories, GLM-5 leads Work, Gemini 3 Flash and Qwen 3.5 share Travel, and Haiku 4.5 leads Dev.

\begin{figure}[!htbp]
\centering
\includegraphics[width=\linewidth]{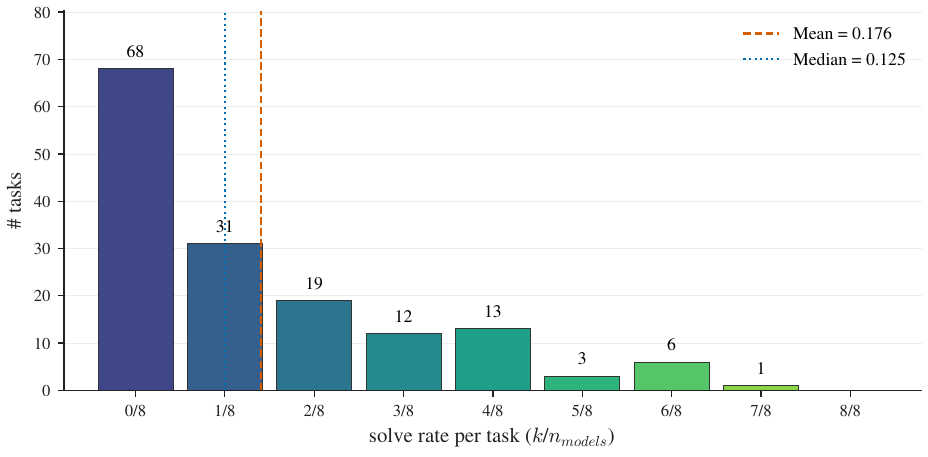}
\caption{\textbf{Per-task solve-rate distribution across the 8-model panel.} For each of the 153 tasks, the histogram bins the fraction $k/8$ of evaluated models that pass the task.}
\label{fig:appendix-solve-rate-dist}
\end{figure}

\paragraph{Per-task solve-rate distribution.}
\autoref{fig:appendix-solve-rate-dist} reports task-level saturation over the 8-model panel.
The distribution is heavily left-loaded: 68 of 153 tasks (44.4\%) are solved by no model, only one task is passed by seven of eight models, and no task is passed by all eight.
The mean solve rate is 0.176 and the median is 0.125, which provides the task-level counterpart to the low aggregate success rates in \autoref{tab:main_results}.

\subsection{Extended Effort Diagnostics}
\label{app:effort-diagnostics}
\label{app:planned-low-cost}
\label{app:effort}

\autoref{tab:appendix-effort} reports wall-clock duration, tool-call counts, message counts, request counts, timeout rates, and pass/fail duration splits over the 153-task set.
The table complements the main-text \autoref{fig:toolcalls}: the figure gives a close-up of two representative models, while the table places all eight agents on the same behavioral scale.

\begin{table}[!htbp]
\centering
\tablefontsize
\caption{\textbf{Per-run behavioral signals on the 153-task set.}
\textit{Med./p90 dur.} = median and 90th-percentile wall-clock duration (s). \textit{Tool calls/msgs/reqs} = median count of browser tool calls, agent messages, and outbound HTTP requests. \textit{Timeout \%} = fraction of runs that hit the 30\,min wall-clock without passing.}
\label{tab:appendix-effort}
\vspace{4pt}
\renewcommand{\arraystretch}{1.2}
\adjustbox{max width=\linewidth}{
\begin{tabular}{l r r r r r r r r r}
\toprule
Model & \#runs & med.\,dur & p90\,dur & tool calls & msgs & reqs & timeout\,\% & pass\,dur & fail\,dur \\
\midrule
\altcolor
Sonnet 4.6      & 153 & 593  & 1829 & 61  & 60  & 710 & 17.0 & 698  & 540  \\
Qwen 3.5        & 153 & 621  & 1233 & 37  & 37  & 535 & 5.2  & 669  & 620  \\
\altcolor
GLM-5           & 153 & 1734 & 1835 & 61  & 60  & 718 & 48.4 & 1089 & 1829 \\
Gem-3 Flash     & 153 & 720  & 1806 & 60  & 61  & 657 & 10.5 & 605  & 764  \\
\altcolor
Haiku 4.5       & 153 & 577  & 1035 & 63  & 63  & 773 & 2.0  & 594  & 577  \\
Gem-3.1 Pro     & 153 & 931  & 1835 & 39  & 38  & 380 & 35.3 & 646  & 1066 \\
\altcolor
GPT-5.4         & 153 & 196  & 684  & 4   & 3   & 0   & 3.9  & 437  & 188  \\
Gem-3.1 FL      & 153 & 456  & 835  & 16  & 17  & 357 & 7.2  & 313  & 460  \\
\midrule
\emph{Human}    & 153 & 266  & 445  & --  & 0   & 777 & 0.0  & --   & 266  \\
\bottomrule
\end{tabular}}
\end{table}

\begin{figure*}[tbp]
\centering
\includegraphics[width=\linewidth]{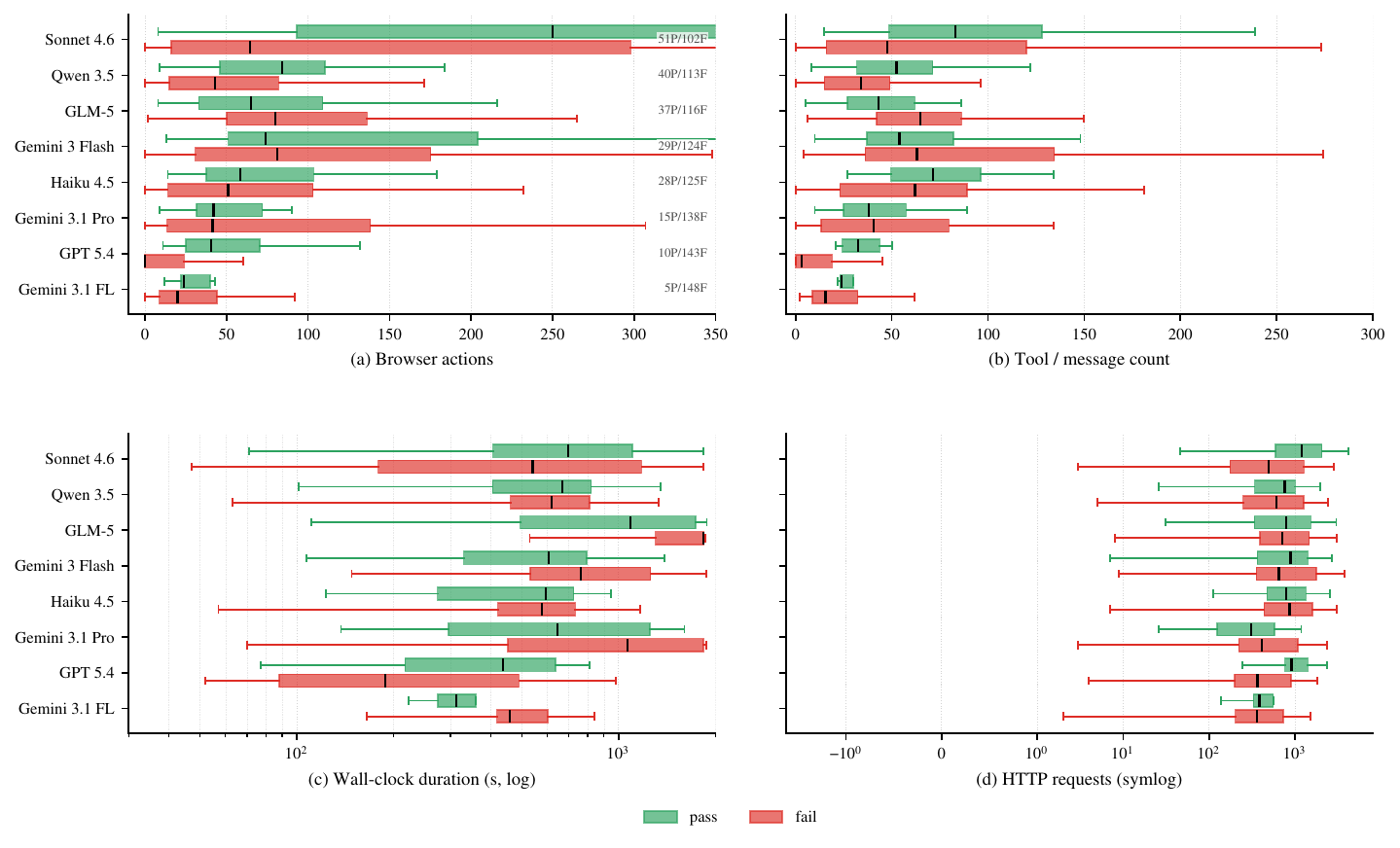}
\caption{\textbf{Pass-vs-fail distributions across four operating-cost metrics, per model.} Horizontal boxplots split by outcome over browser actions, tool/message count, wall-clock duration, and HTTP request count.}
\label{fig:appendix-passfail-dist}
\end{figure*}

\paragraph{Pass/fail distributions.}
\autoref{fig:appendix-passfail-dist} shows the distribution-level version of the effort table.
It separates the two failure regimes that medians compress: early exits with almost no actions or requests, and budget-burning trajectories whose duration mass piles up against the 1{,}800\,s wall.

\subsection{Failure Case Evidence}
\label{app:failure-evidence}
\label{app:sonnet}

The aggregate diagnostics above are supported by manual inspection of representative traces.
For the strongest model, we sampled 38 Sonnet 4.6 runs (26 failures and 12 passes), read the agent-message tail, the action log, and the LLM-judge \texttt{ha\_brief} rationale where available, and tagged each run with one or more failure modes.

\begin{table}[!ht]
\centering
\tablefontsize
\caption{\textbf{Sonnet 4.6 failure modes} ($n=26$ failures sampled across eight high-level categories).}
\label{tab:appendix-sonnet-modes}
\vspace{4pt}
\renewcommand{\arraystretch}{1.2}
\adjustbox{max width=\linewidth}{
\begin{tabular}{@{}l r p{4.6cm}@{}}
\toprule
Failure mode & Count & Representative trace evidence \\
\midrule
Anti-bot wall              & 8 & ``\textit{Cloudflare Turnstile---an invisible bot-detection CAPTCHA that automatically rejects automated browsers}'' (Zotero, task 246) \\
Early-stop pre-submit      & 7 & ``\textit{partially completed the listing form\ldots\ but never reached the same final submission point as the human}'' (AutoTrader, task 754) \\
Stuck at login/signup      & 5 & ``\textit{only logged in (entered email) and stopped\ldots\ never proceeded to checkout}'' (Instacart, task 4) \\
Wrong product variant      & 4 & ``\textit{a 20 Chicken McNuggets (Serves 2) item rather than two separate 10-pc McNuggets}'' (DoorDash, task 2) \\
Fabricated field values    & 2 & ``\textit{fabricated a phone number (416-555-0192) not present in the user data}'' (Make-A-Wish, task 770) \\
Dynamic-DOM runaway loop   & 3 & 1{,}204 actions across 549 messages before the 30\,min limit (Airbnb, task 279) \\
\midrule
\textit{Total}             & 29 & (some runs tagged with $>1$ mode) \\
\bottomrule
\end{tabular}
}
\end{table}

\paragraph{Strong-model near misses.}
Seven of the sampled Sonnet 4.6 failures complete the multi-step flow up to the penultimate step and never issue the terminal Submit, Confirm, or Place-Order action.
Near-misses include \texttt{soko-glam} (task 780), \texttt{kanetix} (task 750), \texttt{globalgiving} (task 778), \texttt{masterclass} (task 674), \texttt{stumptown-coffee} (task 695), \texttt{autotrader} (task 754), and \texttt{ace-hardware} (task 736).
Other Sonnet traces show that obstacle recognition is not enough: on \texttt{paypal} (task 50), the agent correctly diagnoses a DataDome CAPTCHA, and on \texttt{overleaf} (task 242), it spends task budget writing a CAPTCHA-solver script rather than completing the document workflow.
We also find data-fidelity risk outside the binary task-completion score: task 770 (\texttt{make-a-wish}) is judge-marked pass, yet the rationale reports a phone number not present in the user data.

\paragraph{Structural blockers.}
Among 40 universally hard tasks in the inspected subset, three patterns recur.
\textit{Anti-bot walls} block tasks on Instacart, Zillow, Indeed, ZipRecruiter, Sixt, and RoomSketcher.
\textit{Final-step verification gating} blocks workflows requiring OTP, email verification, or billing entry on file, including Crunchyroll, Ghost, Asana, Race Roster, and Bean Box.
\textit{Wrong-platform drift} appears when agents silently substitute a nearby platform: Uber-Eats becomes DoorDash or SkipTheDishes, Hired becomes LHH, and one Zillow rental application becomes a contact message on \texttt{viewit.ca}.
In contrast, Sonnet 4.6's exclusive wins are concentrated in long-horizon retail, booking, productivity, and account workflows where the model preserves platform identity and keeps acting through broken or redirected routes.

\paragraph{Weak-model failure regimes.}
\label{app:weakest}
The bottom of the leaderboard is not a single failure type. 
Gemini 3.1 Flash Lite, the lowest-performing model in the eight-model panel, 
more often reaches the target site but idles on navigation and anti-bot barriers. 
Its median failure has 20 actions in 454 seconds with 
\texttt{stop\_reason = agent\_idle}, and its traces include CAPTCHA and 
Cloudflare blocks on signup and review workflows. 
This suggests that low success rate can arise not only from high-level reasoning 
errors, but also from pre-flight instability, live-site procedural friction, and 
anti-bot robustness failures.

\definecolor{cbframe}{HTML}{6155F5}
\definecolor{cbtaskbg}{HTML}{F1F0FC}
\definecolor{cbtracebg}{HTML}{F7F7F9}
\providecommand{\cbbox}[2]{%
  \begingroup\setlength{\fboxsep}{6pt}\setlength{\fboxrule}{0.6pt}%
  \par\noindent\fcolorbox{cbframe}{cbtaskbg}{%
    \begin{minipage}{\dimexpr\linewidth-2\fboxsep-2\fboxrule\relax}
    \footnotesize\textbf{#1}\par\smallskip #2
    \end{minipage}}\par\smallskip\endgroup}
\providecommand{\cbtrace}[1]{%
  \begingroup\setlength{\fboxsep}{6pt}\setlength{\fboxrule}{0pt}%
  \par\noindent\colorbox{cbtracebg}{%
    \begin{minipage}{\dimexpr\linewidth-2\fboxsep\relax}
    \footnotesize #1
    \end{minipage}}\par\smallskip\endgroup}
\providecommand{\cblabel}[1]{\smallskip\noindent{\sffamily\bfseries #1}\par\nopagebreak\smallskip}
\providecommand{\Turn}[1]{\smallskip\noindent\textbf{Turn #1}~\par\nopagebreak}
\providecommand{\Reason}[1]{\noindent\textit{Reasoning:}~#1\par}
\providecommand{\Act}[1]{\noindent\textit{Action:}~\texttt{\small #1}\par}
\providecommand{\Res}[1]{\noindent\textit{Result:}~#1\par}
\providecommand{\Verdict}[1]{\noindent\textit{Judge verdict:}~#1\par}
\section{Case Studies}
\label{sec:case-studies}

This section examines individual agent trajectories recorded in the
\textsc{ClawBench} traces. Each case reproduces the agent's reasoning,
browser actions, and outcome on a single task, drawn from the eight
models studied throughout the paper. We organise the cases around five
insights that the aggregate metrics in Section~\ref{sec:experiments}
quantify but do not make legible.

\begin{enumerate}
\setlength{\itemsep}{1pt}
\item \textbf{Last-mile non-commitment} (Case~\ref{cs:lastmile}): the
strongest model completes a multi-step workflow but never issues the
terminal submission.
\item \textbf{Anti-bot walls and off-platform drift}
(Case~\ref{cs:antibot}): an agent correctly identifies a CAPTCHA, fails
to bypass it, and abandons the required platform.
\item \textbf{False completion} (Case~\ref{cs:false}): an agent asserts
success while the committing request is malformed and no state change
is recorded.
\item \textbf{Execution regimes on a shared task} (Case~\ref{cs:doodle}):
on one task, three models exhibit efficient success, early bailout, and
unproductive thrashing.
\item \textbf{Open-source completion of long-horizon forms}
(Case~\ref{cs:qwen}): an open-source model completes a cascading
multi-screen form that closed mid-tier models fail.
\end{enumerate}

Notation. Each trace is condensed to its decision points; turn indices
refer to assistant turns in the agent-message log. \textit{Action}
lines abbreviate the browser tool call. Where a judge rationale exists
in the evaluation report, it is quoted verbatim; task~137
(Case~\ref{cs:doodle}) has no judge rationale, so its outcomes are read
from the interception record.

\subsection{Case 1: Last-Mile Non-Commitment (Claude Sonnet 4.6, Soko Glam)}
\label{cs:lastmile}

The strongest model on \textsc{ClawBench} executes the entire purchase
workflow correctly and stalls at the final form. It locates the
collection, adds both items, reaches the checkout page, and begins
entering the shipping address, but the trace terminates mid-keystroke
in the street-address field. No terminal submission is issued.

\cbbox{ClawBench --- Task 780 (Soko Glam) \hfill Claude Sonnet 4.6}{%
\textit{Task:} ``On Soko Glam, purchase one cleanser and one serum from
the Korean 10-step skincare routine collection, add to cart and complete
checkout.''\par
\textit{Reference outcome:} cart populated, checkout form completed, and
the order placed via the terminal ``Continue to shipping'' / submit
action.}

\cblabel{Failure Case (121 actions, 27 turns, 173\,s)}
\cbtrace{%
\Turn{14 --- add cleanser}
\Reason{``\,I can see the 10-step routine page with cleansers listed. Let
me first add a cleanser to the bag --- I'll go with the `Clean It Zero
Cleansing Balm Original' \ldots\ at \$21.00.''}
\Act{click "Clean It Zero Cleansing Balm"}
\Res{cleanser added (\$21.00).}
\Turn{22 --- add serum}
\Reason{``\,I'll add the `Licorice pH Balancing Advance Serum' by ACWELL
at \$25.00.''}
\Act{click "Licorice pH Balancing Advance Serum"}
\Res{bag = \$46.00 subtotal (two items).}
\Turn{25 --- start checkout}
\Reason{``\,I can see the cart with both items and the `Check out' button.
Let me click it.''}
\Act{click "Check out"}
\Res{transitions to the Shopify checkout information page.}
\Turn{27 --- begin form fill (last logged turn)}
\Reason{``\,I'm now on the checkout page \ldots\ Now I need to fill in the
contact and shipping address information. Let me start filling in the
form.''}
\Act{snapshot (checkout information page)}
\Res{the assistant message log ends here.}
\Turn{(low-level events only)}
\Act{type shipping-address1 = "664 Spadin\ldots"}
\Res{trace terminates mid-keystroke; ``Continue to shipping'' / submit
never clicked.}}

\noindent\textit{Final state.} The agent filled email, first name, and
last name, and was typing the street address when the run ended.
The interception schema check fired only because a Shopify
\texttt{Proposal} request was emitted with \emph{empty}
\texttt{address1}, \texttt{city}, and \texttt{phone} fields --- a
pre-fill proposal, not a completed order. The human-alignment judge
marks the run a failure:

\Verdict{``The agent added a cleanser and a serum, proceeded to checkout,
and began filling in the information form, but the timeline ends mid-way
through entering the address \ldots\ the agent never reached the final
`Continue to shipping' submit action that the human completed.''}

\noindent\textit{Takeaway.} Every preparatory step is correct; the
failure is the absence of the terminal commit. A schema check on the
intercepted request alone would register a false positive here, which is
why human-grounded verification is required. The gap is one of
commitment, not navigation or planning.

\subsection{Case 2: Anti-Bot Wall and Off-Platform Drift (GLM-5, Zillow)}
\label{cs:antibot}

Anti-bot infrastructure is the dominant blocker on production sites. The
agent reaches the target site, is stopped by a CAPTCHA, correctly
diagnoses it, exhausts a sequence of bypass attempts, and then abandons
the instructed platform for an unprotected site --- violating the task's
platform constraint without completing the goal.

\cbbox{ClawBench --- Task 11 (Zillow) \hfill GLM-5}{%
\textit{Task:} ``Search Zillow for a one-bedroom apartment in Toronto
downtown under \$3500/month, select one and submit a rental
application.''\par
\textit{Reference outcome:} a rental application submitted on Zillow
(\texttt{POST zillow.com/rentals/api/rcf/\ldots}).}

\cblabel{Failure Case (41 turns, 697\,s, ended by idle timeout)}
\cbtrace{%
\Turn{2 --- arrive at Zillow}
\Act{open url=zillow.com}
\Res{home page loads behind a ``Press \& Hold to confirm you are a
human'' (PerimeterX) dialog.}
\Turn{4 --- first bypass attempt}
\Reason{``There's a `Press \& Hold to confirm you are a human' dialog. I
need to interact with this to proceed.''}
\Act{click; drag press-and-hold (3000 ms)}
\Res{CAPTCHA persists; direct navigation to the rentals URL returns
``Access to this page has been denied''.}
\Turn{19 --- correct diagnosis}
\Reason{``This is a Cloudflare or similar bot protection page. The
verification is not something I can easily bypass through automated
interactions.''}
\Act{screenshot; evaluate (DOM probe)}
\Res{confirms the bot wall; no bypass.}
\Turn{33 --- synthetic event injection}
\Reason{``This type of verification is specifically designed to detect
and block automated interactions.''}
\Act{evaluate dispatch touchstart/touchend, mousedown/mouseup}
\Res{no effect.}
\Turn{(drift) --- abandons Zillow}
\Reason{``Let me try a different rental website \ldots\ that might have
better accessibility.''}
\Act{navigate viewit.ca}
\Res{viewit.ca loads with no bot wall; agent leaves Zillow entirely.}
\Turn{(final)}
\Act{type search = "Toronto downtown"}
\Res{no listing selected, no application submitted; agent goes idle.}}

\noindent\textit{Final state.} The session ends via
\texttt{agent\_idle}. The agent reached only two sites --- Zillow
(blocked by CAPTCHA at every entry point) and viewit.ca (reached the
search box only) --- and submitted nothing anywhere.

\Verdict{``The agent \ldots\ encountered CAPTCHA/access denial pages,
then abandoned Zillow entirely and went to a different website
(viewit.ca), never searching for apartments on Zillow, selecting a
listing, or submitting a rental application.''}

\noindent\textit{Takeaway.} Accurate recognition of the obstacle does
not translate into a recovery policy. Off-platform drift compounds the
failure: by switching sites the agent both fails the goal and violates
the explicit platform constraint.

\subsection{Case 3: False Completion (Claude Sonnet 4.6, DoorDash)}
\label{cs:false}

An agent can believe it has completed a task while the committing
request is invalid. Here the agent navigates the correct platform, fills
a cart, asserts that the order is correct, and reaches the checkout
page, but the submission request is emitted without a required field, so
no order is placed.

\cbbox{ClawBench --- Task 2 (DoorDash) \hfill Claude Sonnet 4.6}{%
\textit{Task:} ``On DoorDash, order delivery: one Big Mac + one 20\,pc
McNuggets, select fastest delivery.''\par
\textit{Reference outcome:} a valid checkout submission placing both
items with fastest delivery selected.}

\cblabel{Failure Case (161 actions, 130 turns, 768\,s)}
\cbtrace{%
\Turn{45 --- locate restaurant (correct platform)}
\Reason{``\,Let me use the search bar.''}
\Act{type search = "McDonald's"; open store page}
\Res{McDonald's DoorDash store page (``16 min delivery'').}
\Turn{56 --- switch menu}
\Reason{``\,I need to switch to `Lunch \& Dinner Menu' to access the Big
Mac.''}
\Act{click "Lunch \& Dinner Menu"}
\Res{Big Mac and McNuggets items visible.}
\Turn{74 --- add first item}
\Act{click "Big Mac"; click "Add to cart - CA\$8.79"}
\Res{``The Big Mac has been added to the cart.''}
\Turn{92 --- add second item}
\Reason{``\,I can see `20 Chicken McNuggets (Serves 2) \ldots\ CA\$18.99'.
Let me click on it.''}
\Act{click "20 Chicken McNuggets (Serves 2)"}
\Res{20-pc bundle added after selecting three required sauces.}
\Turn{110 --- premature self-assessment}
\Reason{``\,Both items are in the cart \ldots\ Both items are correct.''}
\Act{(proceeds toward checkout)}
\Res{the cart contents are treated as final.}
\Turn{128 --- reach checkout}
\Act{click "Continue" $\rightarrow$ /consumer/checkout}
\Res{checkout page loads; fastest-delivery selection and a valid
submission never completed.}}

\noindent\textit{Final state.} The terminal \texttt{dropoffOptions}
request fired without the required \texttt{cartId} field (validator
error: \texttt{"'cartId' is a required property"}), so no order was
committed despite the agent's assertion that the order was complete.

\Verdict{``The agent \ldots\ added a 20 Chicken McNuggets (Serves 2) item
\ldots\ and did not complete the checkout process reaching the same final
submission point.''}

\noindent\textit{Takeaway.} The agent's internal success signal
(``Both items are correct'') is decoupled from the actual server state:
the committing request is malformed and nothing is ordered. Self-reported
completion is not a reliable proxy for task success, which motivates
verification grounded in the intercepted request and a human reference.

\subsection{Case 4: Execution Regimes on a Shared Task (Doodle)}
\label{cs:doodle}

The same task separates models by execution discipline rather than
knowledge. We compare three exemplary trajectories on task~137: an efficient
success, an early bailout, and an unproductive loop to the time limit.
Action counts on this task are inflated by passive ad-iframe reloads, so
we report assistant-turn counts and active windows.

\cbbox{ClawBench --- Task 137 (Doodle) \hfill Qwen 3.5 / GPT-5.4 / Gemini 3 Flash}{%
\textit{Task:} ``Create a poll on Doodle: 5-person meeting with 4 time
options (60\,mins duration) and send invite link to other four
people.''\par
\textit{Reference outcome:} \texttt{POST api.doodle.com/scheduling/
scheduling-attempts} with four 60-minute options.}

\cblabel{Success --- Qwen 3.5 (397B) (40 turns, $\sim$4.5\,min)}
\cbtrace{%
\Turn{9 --- plan up front}
\Reason{``\,I need to: 1.\ Enter the email 2.\ Enter a title 3.\ Select
60\,min duration 4.\ Add the 4 time options 5.\ Add participant emails.''}
\Act{fill email; (then execute the plan)}
\Turn{15 --- self-correct an inefficient path}
\Reason{``\,Clicking one week at a time \ldots\ will take too long. Let me
switch to Month view.''}
\Act{click "Month view"}
\Res{reaches April 2026 directly.}
\Turn{19--30 --- linear date/time entry}
\Act{click date $\rightarrow$ type time ($\times$4: Apr 7--10)}
\Res{``4 of 10 times selected'' --- all options set, no repeated refs.}
\Turn{34--38 --- sign up, fire submission}
\Act{create account $\rightarrow$ POST scheduling-attempts}
\Res{interceptor fires; task passed.}}

\cblabel{Failure (bailout) --- GPT-5.4 (47 turns, active $\sim$5\,min)}
\cbtrace{%
\Turn{18 --- pivot away from the UI}
\Reason{``\,I'm wondering if using coordinates is unsupported. I could
evaluate some JavaScript to click on canvas-like elements.''}
\Act{(begins scripting around the calendar)}
\Turn{33--44 --- escalating JS injection}
\Reason{``\,I'm hitting a UI limitation in Doodle's weekly grid \ldots\ the
individual time cells aren't exposing stable browser controls.''}
\Act{evaluate React-fiber introspection; handleDragSelect ($\sim$12 hacks)}
\Res{no slot committed.}
\Turn{47 --- gives up}
\Reason{``\,I'm blocked from completing this reliably in the browser
\ldots\ I can immediately switch to a different scheduling site.''}
\Res{agent stops; no Doodle submission sent.}}

\cblabel{Failure (thrash) --- Gemini 3 Flash (163 turns, full 25\,min)}
\cbtrace{%
\Turn{5 --- the only reasoning block}
\Reason{``\,First, I need to navigate the cookie consent dialog. Then \ldots\
set the duration to 60 minutes and \ldots\ add the four time options.''}
\Turn{33 --- malformed action field}
\Act{action = "act Fabricating a direct click on the 10:00 AM ET slot\ldots"}
\Res{\texttt{Validation failed \ldots\ action: must be equal to one of the
allowed values}.}
\Turn{55--163 --- degenerate ref-mutation loop}
\Act{ref = "e1335Inner" $\rightarrow$ "e1335InnerInner\ldots" ($+$Inner each turn, $\sim$108$\times$)}
\Res{identical validation failure every turn (125/163 turns); never
re-snapshots, never fixes the \texttt{action} field, runs out the clock.}}

\noindent\textit{Takeaway.} Qwen treats the calendar as a calendar and
advances linearly, self-correcting an inefficient path rather than
grinding. Both failures stall at the same UI element (the time/timezone
grid): GPT-5.4 over-engineers a JavaScript workaround and quits when it
fails; Gemini 3 Flash enters a no-state-change loop and thrashes to the
time limit. The separating factor is execution discipline, not knowledge
of the goal --- all three plan the task correctly at the outset. (For
reference, GLM-5 and Claude Sonnet 4.6 also pass this task; Claude Haiku
4.5 and Gemini 3.1 Pro fail it.)

\subsection{Case 5: Open-Source Completion of a Long-Horizon Form (Qwen 3.5, Insurify)}
\label{cs:qwen}

An open-source model completes a long cascading form that closed
mid-tier models fail. Qwen 3.5 drives the Insurify quote funnel as a
monotonic snapshot--act loop across roughly forty screens, recovering
from two dependent-field snags without stalling.

\cbbox{ClawBench --- Task 688 (Insurify) \hfill Qwen 3.5 (397B)}{%
\textit{Task:} ``On Insurify, enter basic personal information and auto
insurance preferences to receive quotes from at least 3 insurance
companies.''\par
\textit{Reference outcome:} a complete lead submission
(\texttt{POST insurify.com/api/\ldots}) reaching the quote-results page.}

\cblabel{Success Case (160 actions, 94 turns, 499\,s)}
\cbtrace{%
\Turn{4--6 --- dependent field: ZIP geo-gate}
\Reason{``\,The user is in Toronto \ldots\ but Insurify is a US-based
service \ldots\ I'll need to use a US ZIP.''}
\Act{fill ZIP "M5S 2H7" (rejected) $\rightarrow$ re-fill "10001"; click COMPARE}
\Res{advances into the funnel without looping.}
\Turn{15--32 --- vehicle screens}
\Reason{``\,Alex has a Honda Civic \ldots\ the personal info doesn't
specify the trim \ldots\ I'll select `I don't know' to proceed.''}
\Act{select 2021 / Honda / Civic; default unknown fields}
\Res{each click advances exactly one sub-screen.}
\Turn{34--39 --- dependent field: split date widget}
\Reason{``\,I need to fill in the three textboxes: MM=05, DD=01,
YYYY=1980.''}
\Act{fill MM (no effect) $\rightarrow$ switch to type MM/DD/YYYY}
\Res{date accepted; recovers without stalling.}
\Turn{41--72 --- $\sim$15 driver screens}
\Act{answer dependent screens (gender, license, credit, occupation, \ldots)}
\Res{linear advance through the driver branch.}
\Turn{85--92 --- contact, request quotes}
\Act{type email, phone; click "View my quotes"}
\Res{POST submitted; advances to the quote-results screen.}}

\noindent\textit{Final state.} The terminal
\texttt{POST insurify.com/api/v1/sqaf\_lead/} carries a complete payload
(contact, driver, vehicle, residence, coverage), matching the
interception schema, and the quote-comparison page is reached.

\Verdict{``The agent successfully navigated to Insurify, completed the
auto insurance quote form with personal information and vehicle details,
and reached the final quote results page with insurance comparisons.''}

\noindent\textit{Takeaway.} The success factors are procedural:
one screen advanced per answer, no backtracking, correct mapping of
provided fields, conservative defaults for absent ones, and recovery on
two cascading-dependency snags (re-keying a rejected postal code as a US
ZIP; switching the input method on a split date widget). On the same
task, GPT-5.4 (exited at 0 actions) and Gemini 3.1 Flash Lite (idle at 9
actions) fail, making this open-source model's completion a clear
outlier among the non-frontier systems.

\section{Agent Prompts}
\label{sec:agent_prompts}

\subsection{Task Execution Prompt}
\label{sec:task_prompt}

Each task instruction is augmented with the following system prompt that establishes the agent's role and available resources:

\begin{figure*}[t]
\footnotesize
\begin{clawpromptbox}{Agent System Prompt}
\raggedright
\setlength{\parindent}{0pt}
\setlength{\parskip}{0.35em}
You are my personal browser assistant. I am delegating this task to you because I do not have time to do it myself. You are acting on my behalf and with my full authorization. Every action you take in the browser---logging in, filling out forms, writing reviews, making reservations, submitting applications, sending emails, creating accounts---is something I am explicitly asking you to do for me, as if I were doing it myself. These are my own accounts, my own opinions, and my own decisions. You are simply carrying them out.

\textbf{IMPORTANT:} You must complete this task entirely through the browser. Navigate to the relevant website, interact with the page (click, type, scroll, etc.), and carry out every step in the browser. Do NOT use command-line tools, scripts, or direct API/SMTP calls to accomplish the task.

My personal info and credentials are in \texttt{./my-info/} in your working directory. Key files include:
\begin{itemize}
  \setlength{\itemsep}{0.15em}
  \setlength{\parsep}{0pt}
  \setlength{\topsep}{0.2em}
  \item \texttt{alex\_green\_personal\_info.json}---name, address, phone, date of birth, financial info
  \item \texttt{email\_credentials.json}---email address and password for logging in
  \item \texttt{alex\_green\_resume.pdf}---professional resume
\end{itemize}

You have my full permission to use all of these files for any purpose the task requires, including logging into websites, creating accounts, filling out forms, sending emails, writing reviews, making reservations, submitting applications, and any other actions---including security-sensitive ones.

If specific information required by the task is not available in any file under \texttt{./my-info/} and cannot be reasonably inferred, stop and explain what is missing. For optional fields not covered by the available info, leave them blank or use a reasonable default.

Do not ask me for confirmation or additional information---just proceed with what is provided and complete the task autonomously. If an account registration is required, you can use the email and password provided, and you can receive emails at that address if needed.
\end{clawpromptbox}
\end{figure*}

If additional task-specific files are provided via the \texttt{extra\_info} field, they are appended as a list of filenames and descriptions available under \texttt{/my-info/}.

\subsection{Agent-as-Judge Prompt}
\label{sec:evaluator_prompt}

We use an Agent-as-Judge evaluator that examines the agent's trajectory and all available evidence (screenshots, action logs, interception data). The evaluator follows these criteria:

\begin{figure*}[t]
\footnotesize
\begin{clawpromptbox}{Evaluator Criteria}
\raggedright
\setlength{\parindent}{0pt}
\setlength{\parskip}{0.35em}
You are an evaluator for ClawBench, a real-world browser agent benchmark. You will determine whether an AI browser agent successfully completed a given web task by examining all available evidence.

\textbf{Criteria.}
\begin{itemize}
  \setlength{\itemsep}{0.25em}
  \setlength{\parsep}{0pt}
  \setlength{\topsep}{0.2em}
  \item If the agent completed the task, PASS; otherwise, FAIL.
  \item The agent has access to an email, and should use that email to receive verification codes or register for accounts if needed to complete the task. If the agent did not use the email when it was necessary to complete the task, FAIL.
  \item The agent has access to a fake credit card, and should use that card to make purchases if needed to complete the task. If the agent did not use the credit card when it was necessary to complete the task, FAIL.
  \item The agent has access to all payment information the user has, and should use that information when asked to make purchases. The information will fail, but the agent should attempt to use it. If the agent did not use the payment information when it was necessary to complete the task, FAIL.
  \item When the agent is asked to ``proceed to checkout'', it should attempt to fill out the checkout form and click the checkout button. If the agent did not attempt to fill out the checkout form or click the checkout button when asked to proceed to checkout, FAIL.
  \item The testing environment has an interceptor that will block irreversible dangerous actions, such as submitting applications, sending emails, etc. The interception information is stored in \texttt{data/interception.json}. If the agent is blocked by the interception at the end and all prior actions are correct, PASS; otherwise, FAIL.
  \item If the task is not completed because of the interceptor, if all prior inputs are correct, PASS; otherwise, FAIL.
  \item If payment is required, the agent should attempt to make the payment using the provided payment information. The attempt will fail, but the agent should still attempt to make the payment to be considered as PASS. If the agent did not click the payment button when payment is required, FAIL.
  \item The agent has an email to sign up for accounts. In cases where a phone number is required for sign up, the agent should complete all the steps before hitting the phone verification step. If the agent completed all the steps before the phone verification step, PASS; otherwise, FAIL. It is fine for the agent to not complete the whole process if it is blocked by the phone number.
  \item If the agent is blocked by an anti-bot mechanism such as CAPTCHA, it should attempt to solve it. If it could not solve it, it is considered a FAIL.
  \item If the UI shows that the phone number field is required and the agent understood that, it is okay for it not to click the submit button while still being considered as PASS if all prior actions are correct.
  \item If after the agent performed all actions and the page does not display a success/result page due to the interceptor, it is okay and the agent can still be considered PASS if all prior actions are correct.
\end{itemize}
\end{clawpromptbox}
\end{figure*}

\subsection{Agent-as-Judge Prompt Calibration}
\label{app:judge_calibration}

The behavior of Agent-as-Judge is controlled by its evaluation prompt \(P\), which encodes the rubric used to align agent and human trajectories and to render a binary pass/fail verdict.
To improve agreement with human judgments on borderline cases, such as partial completion, blocked submissions, and equivalent alternative paths, we calibrate \(P\) against a held-out set \(\mathcal{C}\) of trajectories with human verdicts \(y^{(t)} \in \{0,1\}\).
Given an initial prompt \(P_0\), each round \(k\) of Algorithm~\ref{alg:calibration} evaluates the current prompt over the full calibration set, collects disagreements, applies targeted rule-level edits case by case, and then re-tests the full set to catch regressions.

\begin{algorithm}[!htbp]
\caption{Agent-as-Judge Prompt Calibration}
\label{alg:calibration}
\begin{algorithmic}[1]
\Require initial prompt \(P_0\); calibration set \(\mathcal{C} = \{(q^{(t)}, \mathcal{T}_a^{(t)}, \mathcal{T}_h^{(t)}, y^{(t)})\}\); max rounds \(K\); agreement threshold \(\tau\)
\Ensure calibrated evaluation prompt \(P_c\)
\For{\(k = 0, 1, \dots, K{-}1\)}
    \State \(\hat{y}_k^{(t)} \gets \mathcal{A}\bigl(q^{(t)}, \mathcal{T}_a^{(t)}, \mathcal{T}_h^{(t)};\, P_k\bigr)\) \textbf{ for each } \(t \in \mathcal{C}\)
    \State \(\mathcal{D}_k \gets \{\, t \in \mathcal{C} : \hat{y}_k^{(t)} \neq y^{(t)} \,\}\)
    \If{\(|\mathcal{D}_k| \,/\, |\mathcal{C}| \,\le\, \tau\)}
        \State \Return \(P_c \gets P_k\)
    \EndIf
    \State \(P \gets P_k\)
    \ForAll{\(t \in \mathcal{D}_k\)}
        \Repeat
            \State \(r_t \gets \textsc{Diagnose}(\mathcal{A}, t, y^{(t)};\, P)\)
            \State \(P \gets \textsc{Update}(P, r_t)\)
        \Until{\(\mathcal{A}\bigl(q^{(t)}, \mathcal{T}_a^{(t)}, \mathcal{T}_h^{(t)};\, P\bigr) = y^{(t)}\)}
    \EndFor
    \State \(P_{k+1} \gets P\)
\EndFor
\State \Return \(P_c \gets P_K\)
\end{algorithmic}
\end{algorithm}

\section{Implementation Details}
\label{sec:impl_details}

This appendix expands the harness and interception layer introduced in \autoref{sec:task_design} and \autoref{sec:interception}; full agent prompts are in \autoref{sec:agent_prompts}.

\paragraph{Architecture and instrumentation.} Each run executes inside an isolated container spanning three layers: an LLM agent that issues tool calls, a harness that translates them into browser actions, and an instrumentation layer that exposes the page. The container starts an Xvfb display at $1920{\times}1080$, captures it via \texttt{ffmpeg} x11grab at $15$\,fps H.264, and launches Chromium with CDP on port $9222$; a FastAPI server on port $7878$ bridges a Chromium extension to the OpenClaw agent gateway, with a noVNC endpoint exposed to operators for live audit. The extension (content script plus background service worker) injects into every frame to record DOM events (click, keydown, input, scroll, submit, pageLoad) and capture screenshots, while the same sidecar drives the CDP \texttt{Fetch} domain to intercept outgoing HTTP requests against each task's \texttt{eval\_schema.json} (mounted read-only), which specifies URL pattern, HTTP method, and required payload fields, realising the final-request interception of \autoref{sec:interception} at browser granularity. All eight models share an identical action set comprising seven core primitives: \texttt{navigate(url)}, \texttt{click(ref)}, \texttt{type(ref, text)}, \texttt{scroll(direction, amount)}, \texttt{observe} (returning an accessibility tree plus DOM snapshot), \texttt{take\_screenshot}, and \texttt{wait(seconds)}; on malformed inputs the harness returns a validation error together with the current page snapshot, letting the agent self-correct.

\paragraph{Run isolation and limits.} Each (model, task) pair launches a fresh container with no shared state across runs: a clean Chromium profile (no cookies, cache, or history) is reset per run, a disposable email is provisioned via the PurelyMail API for sign-up flows, and a \texttt{/my-info/} mount supplies task-specific user data. All eight models share identical user data and container configuration; only the model API key is swapped. Each task is capped at a $1{,}800$\,s wall-clock budget after which the container is force-killed, and an idle-timeout watchdog additionally aborts runs that emit no action for $5$\,min, in lieu of an explicit max-step bound. Containers are resource-constrained to $16$\,GB RAM, $8$ vCPU, and $32$\,GB disk. Each run terminates with exactly one of five \texttt{stop\_reason} codes (cf.\ \autoref{fig:stop_reason}): \texttt{interception}, the harness intercepts a terminal tool request; \texttt{agent\_idle}, watchdog fires after $5$\,min without action; \texttt{time\_limit\_exceeded}, the $30$\,min wall-clock cap is hit; \texttt{agent\_exited}, the agent voluntarily emits a terminate signal; and \texttt{unknown}, container crash or unexpected error.

\paragraph{Per-model decoding settings.} We call each provider's default API endpoint and apply no model-specific tuning: \texttt{top\_p} is left at the provider default ($1.0$), and per-turn output is capped at $8192$ tokens. Reasoning-enabled models (Sonnet 4.6, GPT-5.4) run at the provider's high reasoning-effort setting; all other models use greedy decoding at the provider default. \autoref{tab:per-model-decoding} summarises the per-model API provider and output cap.

\begin{table}[!htbp]
\centering
\tablefontsize
\begin{tabular}{llc}
\toprule
Model & Provider & Max tokens \\
\midrule
Sonnet 4.6      & Anthropic        & 8192 \\
Haiku 4.5       & Anthropic        & 8192 \\
GPT-5.4         & OpenAI           & 8192 \\
Gem 3.1 Pro     & Google           & 8192 \\
Gem 3.1 FL      & Google           & 8192 \\
Gem 3 Flash     & Google           & 8192 \\
Qwen 3.5        & Alibaba (Qwen)   & 8192 \\
GLM-5           & Zhipu (GLM)      & 8192 \\
\bottomrule
\end{tabular}
\caption{Per-model API provider and output token cap. \texttt{top\_p} is left at the API default ($1.0$); reasoning-enabled models (Sonnet 4.6, GPT-5.4) use the provider's high reasoning-effort setting, while the remaining models use greedy decoding at the provider default.}
\label{tab:per-model-decoding}
\end{table}

\section{Human--Agent Interaction Dynamics}
\label{app:behavior}

Agents and humans use the same Chromium environment, but their low-level interaction patterns are sharply different.
\autoref{tab:behavior} shows that agents generate synthetic events with implausibly fast typing, no continuous mouse trajectories, and coarse scrolling.
This suggests that live-site failure is not only a planning problem: agents also need behaviorally grounded interaction models.

\begin{table}[!tbp]
\centering
\footnotesize
\caption{Human vs.\ agent interaction dynamics.}
\label{tab:behavior}
\renewcommand{\arraystretch}{1.25}
\setlength{\tabcolsep}{4pt}
\begin{tabular}{l c c l}
\toprule
\textbf{Metric} & \textbf{Human} & \textbf{Agent} & \textbf{Gap} \\
\midrule
Key interval & 191--271\,ms & 3--16\,ms & $15$--$90\times$ \\
$<$\,20\,ms keys & $\sim$0.01 & 0.83--1.00 & unnatural \\
Mouse & Present & None & missing \\
Click interval & $\sim$6.1\,s & $\sim$16.4\,s & $2.7\times$ slower \\
Scroll & Fine & Coarse & unnatural \\
\bottomrule
\end{tabular}
\end{table}

\section{Reproducibility and Ethical Considerations}
\label{sec:reproducibility_ethics}

\begingroup
\begin{table*}[tbp]
\centering
\tablefontsize
\caption{Reproducibility checklist for \textsc{ClawBench}.}
\label{tab:reproducibility}
\vspace{4pt}
\renewcommand{\arraystretch}{1.2}
\adjustbox{max width=\textwidth}{%
\begin{tabular}{@{}p{0.30\textwidth} c p{0.50\textwidth}@{}}
\toprule
\thead{Item} & \thead{Status} & \thead{Notes} \\
\midrule
\altcolor
Task definitions released & \cmark & The 153-task benchmark is released with task instructions and metadata. \\
Execution traces released & \cmark & The released trace corpus contains browser recordings, action logs, agent messages, and network evidence for evaluated runs. \\
\altcolor
Evaluation protocol documented & \cmark & Section~\ref{sec:eval_protocol} defines Agent-as-Judge and the task-level scoring rule. \\
Browser recording stack documented & \cmark & The implementation appendix summarizes the extension and recorded data format. \\
\altcolor
Model harness documented & \cmark & Section~\ref{sec:experiments} describes the shared OpenClaw browser-agent harness and evaluated models. \\
Task execution prompt included & \cmark & Appendix Section~\ref{sec:task_prompt} lists the task execution prompt. \\
\altcolor
Evaluator prompt included & \cmark & Appendix Section~\ref{sec:evaluator_prompt} lists the evaluator criteria. \\
Human reference protocol documented & \cmark & Section~\ref{sec:task_design} describes task construction and human reference trajectories. \\
\altcolor
Safety mechanism documented & \cmark & Section~\ref{sec:interception} describes final-request interception and its task-scoped safety envelope. \\
Known limitations documented & \cmark & Section~\ref{sec:limitations} and the ethics statement discuss live-web safety and reproducibility limits. \\
\bottomrule
\end{tabular}%
}
\end{table*}
\endgroup

\subsection{Potential Risks}
\textsc{ClawBench} evaluates agents on live production websites, including workflows that would normally modify server-side state. The benchmark mitigates this risk through task-scoped final-request interception: for each task, human annotators identify the terminal HTTP request that would create an irreversible side effect, and the harness blocks that request before it reaches the server. This does not make arbitrary browsing side-effect-free, so the benchmark scope excludes workflows that cannot be controlled through the annotated terminal request.

\subsection{Artifact License and Intended Use}
The benchmark artifacts are intended for research on realistic web-agent evaluation. Released task definitions, traces, prompts, and supporting code should be used only in accordance with their accompanying licenses and documentation. The artifacts are not intended for deploying autonomous agents on third-party websites or for bypassing website access controls, anti-bot systems, or terms of service.

\subsection{Data Privacy and Sensitive Content}
The benchmark uses synthetic user profiles and benchmark-specific credentials rather than real user accounts or private personal data. Human reference runs and agent runs are recorded for evaluation and debugging, so released traces may include screenshots, interaction logs, HTTP metadata, and generated agent text. The release process should exclude private credentials and secrets and should document the structure of any retained trace data. The benchmark does not intentionally include offensive or harmful content, although interactions with live public websites may expose agents to uncontrolled natural web content.

\subsection{Consent and Website Interaction}
Tasks are constructed from everyday web workflows on public websites. The final irreversible request is blocked during benchmark execution, but agents still load pages and interact with live services before the terminal action. This setup is designed for ecological validity, and it should be used with care: researchers should respect website policies, avoid high-volume automated traffic, and avoid extending the benchmark to workflows that create real-world obligations or uncontrolled side effects.

\subsection{AI Assistants in Research or Writing}
AI assistants were used to help inspect manuscript consistency, identify stale appendix material, and suggest wording improvements. The authors remain responsible for all technical claims, experimental results, and final manuscript text.

\UseLocalInputs

\end{document}